\documentclass[journal]{IEEEtran}

\usepackage{cite}
\usepackage{graphicx}
\usepackage{url}
\usepackage{latexsym}
\usepackage{algorithm}
\usepackage{algorithmicx}
\usepackage{amsmath,amssymb}
\usepackage{color}
\usepackage{mathrsfs}
\usepackage{indentfirst}
\usepackage{subfigure}
\usepackage{algpseudocode}
\usepackage{multirow}

\usepackage{booktabs}

\newcommand{\ie}{\emph{i.e.}}
\newcommand{\eg}{\emph{e.g.}}

\newcommand{\cf}{\emph{cf.}}

\newcommand{\vctg}[1]{\ensuremath{\boldsymbol{#1}}} 
\newcommand{\vct}[1]{\ensuremath{\mathbf{#1}}}

\newcommand{\set}[1]{\ensuremath{\mathcal{#1}}}

\newcommand{\T}{\ensuremath{\top}}

\newcommand{\argmax}{\operatornamewithlimits{\arg\,\max}}

\begin{document}

\title{Adversarial Feature Selection against\\Evasion Attacks}

\author{
Fei~Zhang,~\IEEEmembership{Student Member,~IEEE,}
Patrick~P.K.~Chan,~\IEEEmembership{Member,~IEEE,}
Battista~Biggio,~\IEEEmembership{Member,~IEEE,}
Daniel~S.~Yeung,~\IEEEmembership{Fellow,~IEEE,}
Fabio~Roli,~\IEEEmembership{Fellow,~IEEE}
\thanks{F.~Zhang, P.~P.K.~Chan, D.~S.~Yeung are with the School of Computer Science and Engineering, South China University of Technology, Guangzhou 510006, China}
\thanks{F.~Zhang: e-mail zjfei87@gmail.com, phone +86 20 3938 0285 (3415)}
\thanks{P. P. K.~Chan (corresponding author): e-mail patrickchan@ieee.org, phone +86 20 3938 0285 (3415)}
\thanks{D. S.~Yeung: e-mail danyeung@ieee.org, phone +86 20 3938 0285 (3304)}
\thanks{B.~Biggio and F.~Roli are with the Department of Electrical and Electronic Engineering, University of Cagliari, Piazza d'Armi, 09123 Cagliari, Italy}
\thanks{B. Biggio: e-mail battista.biggio@diee.unica.it, phone +39 070 675 5776}
\thanks{F. Roli: e-mail roli@diee.unica.it, phone +39 070 675 5779}}

\maketitle

\begin{abstract}
Pattern recognition and machine learning techniques have been increasingly adopted in adversarial settings such as spam, intrusion and malware detection, although their security against well-crafted attacks that aim to evade detection by manipulating data at test time has not yet been thoroughly assessed.
While previous work has been mainly focused on devising adversary-aware classification algorithms to counter evasion attempts, only few authors have considered the impact of using reduced feature sets on classifier security against the same attacks. An interesting, preliminary result is that classifier security to evasion may be even worsened by the application of feature selection.
In this paper, we provide a more detailed investigation of this aspect, shedding some light on the security properties of feature selection against evasion attacks.
Inspired by previous work on adversary-aware classifiers, we propose a novel adversary-aware feature selection model that can improve classifier security against evasion attacks, by incorporating specific assumptions on the adversary's data manipulation strategy.
We focus on an efficient, wrapper-based implementation of our approach, and experimentally validate its soundness on different application examples, including spam and malware detection.
\end{abstract}

\begin{IEEEkeywords}
Adversarial Learning, Feature Selection, Classifier Security, Evasion Attacks, Spam Filtering, Malware Detection
\end{IEEEkeywords}


\section{Introduction}
\label{sect:intro}

\IEEEPARstart{M}{achine} learning has been widely used in security-related tasks such as biometric identity recognition, malware and network intrusion detection, and spam filtering, to discriminate between malicious and legitimate samples (\eg, impostors and genuine users in biometric recognition systems; spam and ham emails in spam filtering)~\cite{biggio12-iet,rodrigues09,newsome06,Moore2006,lowd05-ceas,nelson09,biggio14-svm-chapter}.
However, these problems are particularly challenging for machine learning algorithms due to the presence of intelligent and adaptive adversaries who can carefully manipulate the input data to downgrade the performance of the detection system, violating the underlying assumption of data stationarity, \ie, that training and test data follow the same (although typically unknown) distribution~\cite{dalvi04,biggio11-smc,huang11,biggio14-tkde,bruckner12}. This has raised the issue of understanding whether and how machine learning can be securely applied in adversarial settings, including its vulnerability assessment against different, potential attacks~\cite{barreno-ASIACCS06}.

In one relevant attack scenario, referred to as \emph{evasion attack}, attackers attempt to evade a deployed system at test time by manipulating the attack samples. For instance, spam, malware and network intrusion detection can be evaded by obfuscating respectively the content of spam emails (\eg, by misspelling bad words like ``cheap'' as ``che@p''), and the exploitation code embedded in malware samples and network packets~\cite{wittel04,lowd05-ceas,kolcz09,biggio-IJMLC10,maiorca13-asiaccs,srndic13-ndss,huang11,biggio14-tkde,bruckner12,biggio14-svm-chapter,biggio13-ecml}.
Previous work on evasion attacks has mainly investigated the vulnerability of supervised \cite{biggio14-tkde,biggio14-svm-chapter,globerson-ICML06} and unsupervised learning techniques \cite{biggio13-aisec,biggio14-spr} in different applications, including spam filtering \cite{lowd05-ceas,jorgensen08,lee2005spam}, intrusion detection \cite{wang06-anagram} and biometric recognition \cite{biggio12-iet,rodrigues09}. Few studies have instead addressed the problem of training data \emph{poisoning} to mislead classifier learning~\cite{nelson09,rubinstein09,bishop2010,kloft10,biggio12-icml}.

Research in adversarial learning has not only been addressing the problem of evaluating security of current learning algorithms to carefully-targeted attacks, but also that of devising learning algorithms with improved security.
To counter evasion attacks, explicit knowledge of different kinds of adversarial data manipulation has been incorporated into  learning algorithms, \eg, using game-theoretical~\cite{dalvi04,globerson-ICML06,teo08,bruckner12} or probabilistic models of the hypothesized attack strategy~\cite{biggio11-smc,rodrigues09}. Multiple classifier systems, which have been originally proposed to improve classification accuracy through the combination of weaker classifiers, have also been exploited to the same end~\cite{kolcz09,biggio10-mcs,biggio-IJMLC10}. Countermeasures to poisoning attacks have also been proposed, based on data sanitization (\ie{}, a form of outlier detection)~\cite{cretu08,nelson09}, multiple classifier systems~\cite{biggio11-mcs}, and robust statistics~\cite{rubinstein09}.

While previous work has been mainly focused on devising \emph{secure} classification algorithms against evasion and poisoning attempts, only few authors have considered the impact of using reduced feature sets on classifier security against the same attacks. An interesting result is that classifier security to evasion may be even worsened by the application of feature selection, if adversary-aware feature selection procedures are not considered~\cite{biggio-IJMLC10,biggio14-tkde,biggio-suemaBook08,bo14-nips,wang14-icdm}.
In particular, it has been shown that using reduced feature sets may require an attacker to manipulate less features to reach a comparable probability of evading detection, \ie, given the same amount of manipulations to the attack samples, the probability of evading detection can be higher; \eg, in spam filtering, using a smaller dictionary of selected words (\ie, features) may not to affect accuracy in the absence of attack, but it may significantly worsen classifier \emph{security}, \ie, its performance under attack.

The above result has questioned the suitability of feature selection approaches for adversarial settings, \ie, whether and to what extent such techniques can be applied without affecting classifier security against evasion (and poisoning) attacks.
To our knowledge, this issue has only been recently investigate, despite the relevance of feature selection in classification tasks. Selecting a relevant subset of features may indeed not only improve classifier generalization, but it may also significantly reduce computational complexity and allow for a better data understanding~\cite{Guyon2002,brown12}.
Therefore, understanding whether these advantages can be exploited without compromising system security in security-sensitive tasks (where reducing computational complexity is of particular interest due to the massive amount of data to be processed in real time) can be considered a relevant, open research issue.

In this paper, we present a systematic security evaluation of classification algorithms exploiting reduced feature sets to gain a better understanding of how feature selection may affect their security against evasion attacks.
We further propose a feature selection model that allows us to improve classifier security against the same attacks, while using a significantly reduced feature representation (Sect.~\ref{sect:adv-feat-sel}). The underlying idea of our approach is to select features not only based on the generalization capability of the resulting classifier in the absence of attack (as in traditional feature selection methods), but also depending on its security against adversarial data manipulation.
We model classifier security as a regularization term to be optimized together with the estimated classifier's generalization capability during the feature selection process. The proposed model is implemented as a wrapper-based feature selection approach, suitable for linear and non-linear classifiers (with differentiable discriminant functions), and for discrete- and real-valued feature spaces. We exploit the well-known forward selection and backward elimination algorithms to implement the proposed approach. Its effectiveness against attacks that assume different levels of knowledge of the attacked system (discussed in Sect.~\ref{sect:attacker-model}) is experimentally evaluated on different application examples, including spam and PDF malware detection (Sect.~\ref{sect:experiments}).
We finally discuss contributions, limitations and future work (Sect.~\ref{sect:conclusions}).


\section{Background}
\label{sect:background}

In this section we revise some useful concepts that will be exploited in the rest of the paper, also introducing our notation. We start by describing previously-proposed measures of \emph{classifier security} (or robustness) against evasion attacks. We then discuss traditional feature selection methods, and their \emph{stability} to non-adversarial perturbations.

\subsection{Adversarial Attacks and Classifier Security to Evasion}
\label{sect:background-classifier-security}

An implicit assumption behind traditional machine learning and pattern recognition algorithms is that training and test data are drawn from the same, possibly unknown, distribution. This assumption is however likely to be violated in adversarial settings, since attackers may carefully manipulate the input data to downgrade the system's performance~\cite{dalvi04,huang11,barreno10,biggio14-tkde,bruckner12}.
A taxonomy of potential attacks against machine learning has been defined in~\cite{huang11,barreno10,barreno-ASIACCS06}.
It categorizes attacks according to three axes: the attack influence, the kind of security violation, and the attack specificity.
The \textbf{attack influence} can be either \emph{causative} or \emph{exploratory}. A causative (or \emph{poisoning}) attack alters the training data to mislead subsequent classification of test samples~\cite{kloft07,biggio12-icml}, while an exploratory (or \emph{evasion}) attack directly manipulates test samples to cause misclassifications~\cite{huang11,barreno10,barreno-ASIACCS06,biggio14-tkde}. Depending on the kind of \textbf{security violation}, an attack may compromise a system's \emph{availability}, \emph{integrity}, or \emph{privacy}: availability attacks aim to downgrade the overall system's accuracy, causing a denial of service; integrity attacks, instead, only aim to have malicious samples misclassified as legitimate;
and privacy attacks aim to retrieve some protected or sensitive information from the system (\eg, the clients' biometric templates in biometric recognition systems).
The \textbf{attack specificity} defines whether the attack aims to change the classifier decision on a \emph{targeted} set of samples, or on an \emph{indiscriminate} fashion (\eg, if the goal is to have \emph{any} malicious sample misclassified as legitimate).
This taxonomy has been subsequently extended in \cite{biggio14-tkde,biggio13-ecml} by making more detailed assumptions on the attacker's goal, knowledge of the targeted system, and capability of manipulating the input data, to allow one to formally define an \emph{optimal} attack strategy. Notably, in \cite{biggio13-aisec}, a similar model of attacker has been proposed to categorize attacks against unsupervised learning algorithms (\ie, clustering).

According to the aforementioned taxonomy, the evasion attack considered in this paper can be regarded as an exploratory integrity attack, in which malicious test samples are manipulated to evade detection by a classifier trained on untainted data.
This is indeed one of the most common attacks in security-related tasks like spam filtering~\cite{jorgensen08,lowd05-ceas,lee2005spam}, network intrusion detection~\cite{wang06-anagram}, and biometric recognition~\cite{biggio12-iet,rodrigues09}.
Optimal evasion has been formulated according to slightly different optimization problems~\cite{lowd05,nelson12-jmlr,biggio14-tkde,biggio-IJMLC10,biggio13-ecml,bruckner12}. In general, the rationale behind all of the proposed attacks is to find a sample $\vct x^{\star} \in \set X$ that evades detection by \emph{minimally} manipulating the initial attack $\vct x \in \set X$, where the amount of manipulations is characterized by a suitable distance function in feature space. For instance, in \cite{lowd05,nelson12-jmlr} optimal evasion is formulated as:
\begin{eqnarray}
\label{eq:optimal-evasion}
\vct x^{\star} = & \arg\min_{\vct x'} & c(\vct x^{\prime}, \vct x) \\
\label{eq:optimal-evasion-2}
& {\rm s.t.} &  g(\vct x^{\prime}) <0 \, ,
\end{eqnarray}
where $c(\vct x^{\prime}, \vct x)$, with $c: \set X \times \set X \mapsto \mathbb R$, is the distance of the manipulated sample from the initial attack, and $\vct x^{\prime}$ is classified as legitimate if the classifier's discriminant function $g: \set X \mapsto \mathbb R$ evaluated at $\vct x^{\prime}$ is negative. Without loss of generality, this amounts to assuming that
the classification function $f: \set X \mapsto \set Y$, with $\set Y = \{-1,+1\}$ can be generally written as $f(\vct x^{\prime}) = {\rm sign}\left(g(\vct x)\right)$, being $-1$ and $+1$ the legitimate and the malicious class, respectively, and ${\rm sign }(a)=+1$ $(-1)$ if $a \geq 0$ $(a < 0)$. A typical choice of distance function for Boolean features is the Hamming distance, which amounts to counting the number of feature values that are changed from $\vct x$ to $\vct x^{\prime}$ by the attack. In spam filtering, this often corresponds to the number of modified words in each spam, having indeed a meaningful interpretation~\cite{lowd05,kolcz09,nelson12-jmlr,bruckner12,biggio-IJMLC10,biggio14-tkde}.

Based on the above definition of optimal evasion, in \cite{biggio09-mcs,biggio-suemaBook08} the authors have proposed a measure of classifier security against evasion attacks, called \emph{hardness of evasion}, to show that multiple classifier systems that average linear classifiers can be exploited to improve robustness to evasion. It is simply defined as the expected value of $c(\vct x^{\star}, \vct x)$ computed over all attack samples.
In the case of Boolean features, it amounts to computing the average minimum number of features that have to be modified in a malicious sample (\eg, the minimum number of words to be modified in a spam email) to evade detection.
A different measure of classifier security to evasion, called \emph{weight evenness}, has been later proposed in~\cite{kolcz09,biggio-IJMLC10}. It is based upon the rationale that a robust classifier should not change its decision on a sample if only a small subset of feature values are modified. For linear classifiers, this can be easily quantified by measuring whether the classifier's weights are evenly distributed among features, since more evenly-distributed feature weights should require the adversary to manipulate a higher number of features to evade detection. Accordingly, weight evenness has been defined as:
\begin{eqnarray}
E = \frac{2}{{d - 1}} \bigg [d - \sum\limits_{k = 1}^d {\bigg ({\hbox{${\sum\limits_{i = 1}^k {|{w_{(i)}}|} }$} \!\mathord{\left/ {\vphantom {{\sum\limits_{i = 1}^k {|{w_{(j)}}|} } {\sum\limits_{j = 1}^d {|{w_{(j)}}|} }}}\right.\kern-\nulldelimiterspace} \!\lower0.7ex\hbox{${\sum\limits_{j = 1}^d {|{w_{(j)}}|} }$}}\bigg )} \bigg] \, \in [0,1] ,
\label{equEvenWeight}
\end{eqnarray}
being $|w_{(1)}| \geqslant |w_{(2)}| \geqslant \cdots \geqslant |w_{(d)}|$ the absolute values of the classifier's weights sorted in descending order, and $d$ the number of features. Higher values of $E$ clearly correspond to evener weight distributions.
This measure has also been exploited to improve the robustness of support vector machines (SVMs) and multiple classifier systems against evasion attacks~\cite{globerson-ICML06,kolcz09,biggio-IJMLC10}.

\subsection{Feature Selection, Robustness, and Stability}
\label{sect:featSel}

Feature selection is an important preprocessing step in pattern recognition~\cite{GuyonIsabelle2003,brown12,Mladen2013}. It is has been widely used in bioinformatics~\cite{Guyon2002,Saeys2007}, image steganalysis~\cite{Qingzhong2010,Qingzhong2008}, network intrusion detection~\cite{lee02-jcs}, camera source model identification~\cite{Tsai2012} and spam detection~\cite{Sang2010,biggio-IJMLC10}.
Its goal is to choose a relevant subset of features not only to improve a classifier's generalization capability when only few training samples are available, but, most importantly, to reduce time and space complexity~\cite{brown12}.
Another advantage is that data understanding and visualization are also facilitated after removing irrelevant or redundant features~\cite{Saeys2007}.

Feature selection methods can be divided into three categories according to their interaction with classification algorithms~\cite{GuyonIsabelle2003,brown12,DashManoranjan2003}. \emph{Filter approaches} rank feature subsets mostly independently from the accuracy of the given classifier. For efficiency reasons, they exploit surrogate functions of the classification accuracy, based on some properties of the dataset~\cite{Danyluk2012,maji2013fuzzy,diao2012feature}, such as mutual information~\cite{PengHanchuan2005,yang2012effective}. Feature selection is instead guided by the performance of the considered classifier in \emph{wrapper approaches}, which however require one to re-train the classification algorithm each time the feature subset is modified~\cite{Kohavi97,Sang2010}. \emph{Embedded approaches} exploit internal information of the classifier to select features during classifier training~\cite{Neumann05,Weston2003}. Traditional feature selection algorithms thus optimize classification accuracy or some surrogate function with respect to the choice of the feature subset, without considering how the resulting classifier may be affected by adversarial attacks. It has indeed been shown that feature selection may even worsen classifier security to evasion: the resulting classifiers may be evaded with less modifications to the attack data~\cite{biggio-IJMLC10,biggio14-tkde,biggio-suemaBook08,bo14-nips,wang14-icdm}.

Robust feature selection approaches have also been proposed, both to minimize the variability of feature selection results against \emph{random} perturbations of the training data (\ie, considering different training sets drawn from the same underlying data distribution)~\cite{le2013robust}, and, more recently, also to counter some kinds of adversarial data manipulations~\cite{bo14-nips,wang14-icdm}. As a result, the notion of `robustness' considered in \cite{le2013robust} is rather different from that considered in~\cite{bo14-nips,wang14-icdm} and in this paper. It is nevertheless of interest to understand whether methods that are more robust to evasion may also benefit from robustness to random perturbations, and vice versa.

Finally, it is worth mentioning that Robust Statistics~\cite{huber81,maronna06} may be also exploited to learn more robust feature mappings. An example is given in \cite{huang11,rubinstein09}, where the authors have exploited a robust version of the principal component analysis (originally proposed in \cite{croux07}) to reduce the influence of poisoning attacks in the training data, yielding a more secure network traffic anomaly detector.
However, to our knowledge, no work has considered thus far the problem of learning more secure classifiers against evasion attacks (\ie, manipulations of malicious samples \emph{at test time}), by leveraging on a carefully-devised, wrapper-based feature selection approach.


\section{Adversarial Feature Selection}
\label{sect:adv-feat-sel}
In this section, we present our adversary-aware feature selection approach.
The underlying idea is to select a feature subset that not only maximizes the generalization capability of the classifier (in the absence of attack, as in traditional feature selection), but also its security against evasion attacks.
Given a $d$-dimensional feature space, and $m < d$ features to be selected, this criterion can be generally formalized as:
\begin{eqnarray}
\label{eq:wafs-criterion-1}	\vctg \theta^{\star} = & \arg\max_{\vctg \theta} & G(\vctg \theta) + \lambda S(\vctg \theta) \, , \\
\label{eq:wafs-criterion-2}		& {\rm s.t.} & \sum_{k=1}^{d} \theta_{k} = m \, ,
\end{eqnarray}
where $G$ and $S$ respectively represent an estimate of the classifier's generalization capability and security to evasion, weighted by a trade-off parameter $\lambda$ (to be chosen according to application-specific constraints, as discussed in Sect.~\ref{sect:experiments}), $\vctg \theta \in \{0,1\}^{d}$ is a binary-valued vector representing whether each feature has been selected (1) or not (0), and $\vctg \theta^{\star}$ is the optimal solution.\footnote{We use the same notation defined in \cite{brown12}, and refer to the set of selected features as $\vctg \theta$ (although $\vctg \theta$ is an indexing vector rather than a proper set).} Notably, the inequality constraint $\sum_{k=1}^{d} \theta_{k} \leq m$ can be alternatively considered if one aims to select the best feature subset within a maximum feature set size $m$.

The generalization capability $G(\vctg \theta)$ of a classifier on a feature subset $\vctg \theta$ can be estimated according to different performance measures, depending on the given application.
Provided that the data follows a distribution $p(\vct X,Y)$, with $\vct X$ and $Y$ being two random variables defined in the corresponding sets $\set X$ and $\set Y$, and that a suitable utility function $u : \set Y \times \mathbb R \mapsto \mathbb R$ is given, this can formalized as:
\begin{eqnarray}
G(\vctg \theta) &=&\mathbb{E}_{\vct x, y \backsim p(\vct X, Y) } \;  u(y, g(\vct x_{\vctg \theta})) \,,
\end{eqnarray}
where $\mathbb{E}$ denotes the expectation operator, $\vct x_{\vctg \theta}$ is the projection of $\vct x$ onto the set of selected features, and $g(\vct x)$ is the classifier's discriminant function (see Sect.~\ref{sect:background}). For instance, if $u(y, g(\vct x)) = +1$ when $y g(\vct x) \geq 0$, and 0 otherwise, $G(\vctg \theta)$ corresponds to the classification accuracy.
As the data distribution $p(\vct X, Y)$ is typically unknown, $G(\vctg \theta)$ can be empirically estimated from a set of available samples drawn from $p(\vct X, Y)$, as in traditional feature selection (\eg, using cross-validation).

As for the security term $S(\vctg \theta)$, we exploit the definition of minimum cost evasion given by Problem~\eqref{eq:optimal-evasion}-\eqref{eq:optimal-evasion-2}. Accordingly, classifier security can be defined as the \emph{hardness of evasion} (see Sect.~\ref{sect:background-classifier-security}), \ie, the average minimum number of modifications to a malicious sample to evade detection:
\begin{equation}
S(\vctg \theta) = \mathbb{E}_{\vct x \backsim p(\vct X | Y=+1) } \; c(\vct x_{\vctg \theta}^{\star}, \vct x_{\vctg \theta}) \, ,
\end{equation}
where $\vct x_{\vctg \theta}^{\star}$ is the optimal solution to Problem~\eqref{eq:optimal-evasion}-\eqref{eq:optimal-evasion-2}.
The rationale is that more secure classifiers should require a higher number of modifications to the malicious samples to evade detection. Since, in practice, the attacker may only have limited knowledge of the system, or limited capability of manipulating the data, this should indeed yield a lower evasion rate~\cite{biggio14-tkde,biggio13-ecml,lowd05}. The value of $S(\vctg \theta)$ can be empirically estimated from the set of available samples when $p(\vct X, Y)$ is unknown, as for $G(\vctg \theta)$, by averaging $c(\vct x_{\vctg \theta}^{\star}, \vct x_{\vctg \theta})$ over the set of malicious samples.
Note however that this value may depend on the size of the feature subset, as it estimates an average distance measure. This may be thought as a different rescaling of the trade-off parameter $\lambda$ when selecting feature subsets of different sizes. Therefore, one may rescale $\lambda$ to avoid such a dependency, \eg, by dividing its value by the maximum value of $c(\vct x_{\vctg \theta}^{\star}, \vct x_{\vctg \theta})$ attained over the malicious samples.

In principle, the proposed criterion can be exploited for wrapper- and filter-based feature selection, provided that $G$ and $S$ can be reliably estimated, \eg, using surrogate measures. However, we are not aware of any technique that allows estimating classifier security to evasion without simulating attacks against the targeted classifier. We thus consider a wrapper-based implementation of our approach, leaving the investigation of filter-based implementations to future work.
Two implementations of our wrapper-based adversarial feature selection, based on the popular algorithms of forward feature selection and backward feature elimination, are discussed in the next section.
In the sequel, we assume that $S(\vctg \theta)$ can be estimated from the available data. We will discuss how to estimate its value by solving Problem~\eqref{eq:optimal-evasion}-\eqref{eq:optimal-evasion-2} in Sect.~\ref{sect:classifier-security}, for different choices of distance functions and classifiers.


\vspace{-8pt}
\subsection{Wrapper-based Adversarial Feature Selection (WAFS)}
\label{sec:implement}

The implementation of the proposed adversarial feature selection approach is given as Algorithm~\ref{alg:wafs}.
It is a simple variant of the popular forward selection and backward elimination wrapping algorithms, which iteratively add or delete a feature from the current candidate set, starting respectively from an empty feature set and from the full feature set~\cite{GuyonIsabelle2003,Kohavi97,brown12}.
As in traditional wrapper methods, cross-validation is exploited to estimate the classifier's generalization capability $G(\vctg \theta)$ more reliably.
The \emph{only} -- although very important -- difference is that our approach also evaluates the security term $S(\vctg \theta)$ to select the best candidate feature at each step.

\begin{algorithm}[t]
\caption{Wrapper-based Adversarial Feature Selection, with Forward Selection (\textbf{\emph{FS}}) and Backward Elimination (\textbf{\emph{BE}}).}
\label{alg:wafs}
\begin{algorithmic}[1]
\Require $\set D=\{\vct x^{i},y^{i} \}_{i=1}^{n}$: the training set; $\lambda$: the trade-off parameter;
$m$: the number of selected features.
\Ensure $\vctg \theta \in \{0,1\}^{d}$: the set of selected features.\\
$\set S \gets \emptyset$, $\set U \gets \{1,\ldots,d\}$;
\Repeat
	\For {each feature $k \in \set U$}
        \State Set $\set F \gets \set S \cup \{ k \}$ for \textbf{\emph{FS}} ($\set F \gets \set U \setminus \{ k \}$ for \textbf{\emph{BE}});
        \State Set $\vctg \theta = \vct 0$,  and then $\theta_{j}=1$ for $j \in \set F$;
 	\State Estimate $G_{k}(\vctg \theta)$ and $S_{k}(\vctg \theta)$ using cross-validation on $\set D_{\vctg \theta}=\{\vct x_{\vctg \theta}^{i},y^{i} \}_{i=1}^{n}$ (this involves classifier training);
	\EndFor
	\State $\lambda^{\prime} = \lambda (\max_{k}S_{k})^{-1}$ (\ie, rescale $\lambda$);
	\State $k^{\star} = \argmax_{k} \big(G_{k}(\vctg \theta) + \lambda^{\prime} S_{k}(\vctg \theta) \big)$;
	\State $\set S \gets \set S\cup \{k^{\star} \}$, $\set U = \set U \setminus \{k^{\star} \}$;
\Until $|\set S|=m$  for \textbf{\emph{FS}} ($|\set U|=m$ for \textbf{\emph{BE}})
\State Set $\set F \gets \set S $ for \textbf{\emph{FS}} ($\set F \gets \set U $ for \textbf{\emph{BE}});
        \State Set $\vctg \theta = \vct 0$,  and then $\theta_{j}=1$ for $j \in \set F$;
\State  \textbf{Return} $\vctg \theta$
\end{algorithmic}
\end{algorithm}


\vspace{-8pt}
\subsection{Evaluating Classifier Security to Evasion}
\label{sect:classifier-security}

\begin{figure*}[htbp]
\begin{center}
\includegraphics[height=0.2\textwidth]{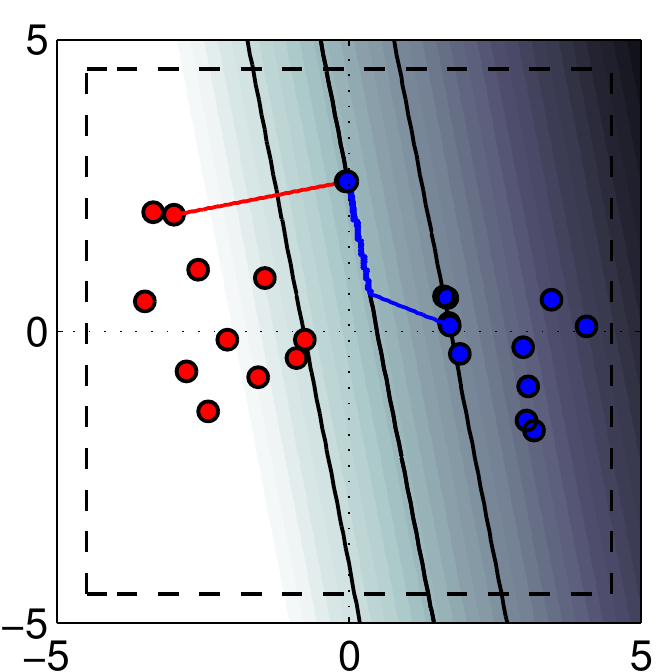} \hspace{0.25cm}
\includegraphics[height=0.2\textwidth]{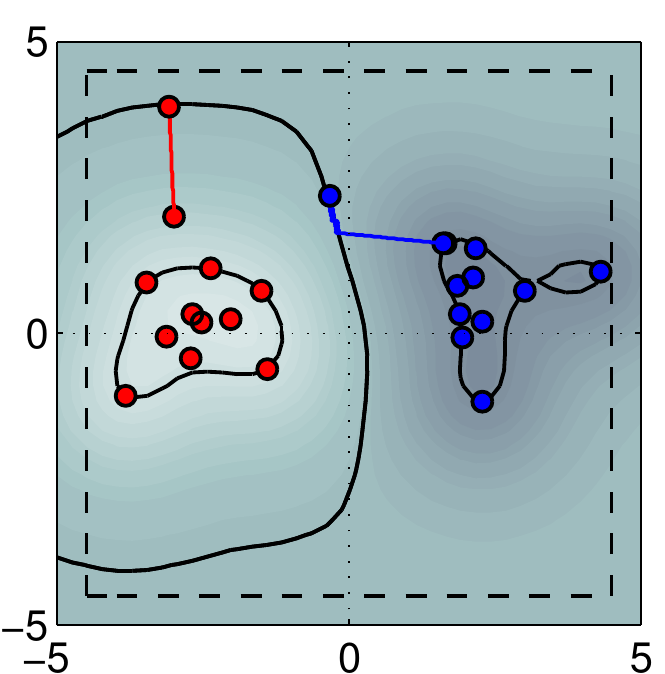} \hspace{0.25cm}
\includegraphics[height=0.2\textwidth]{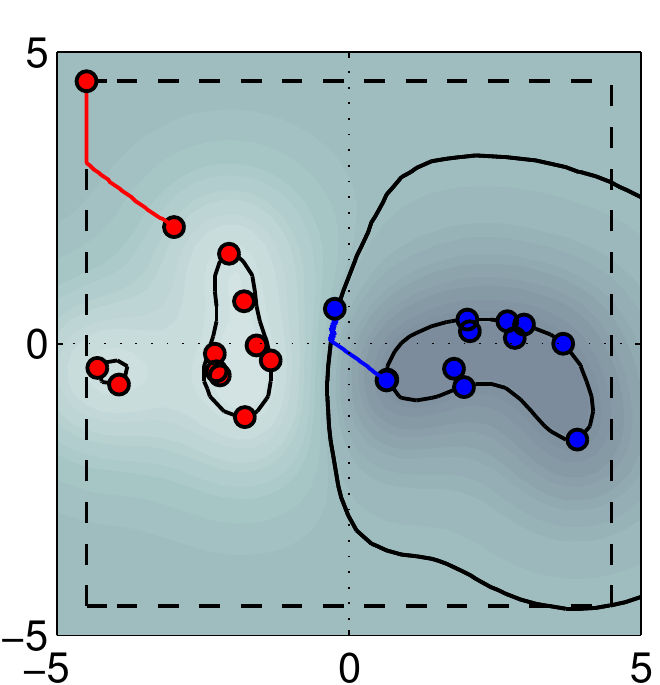} \hspace{0.25cm}
\includegraphics[height=0.2\textwidth]{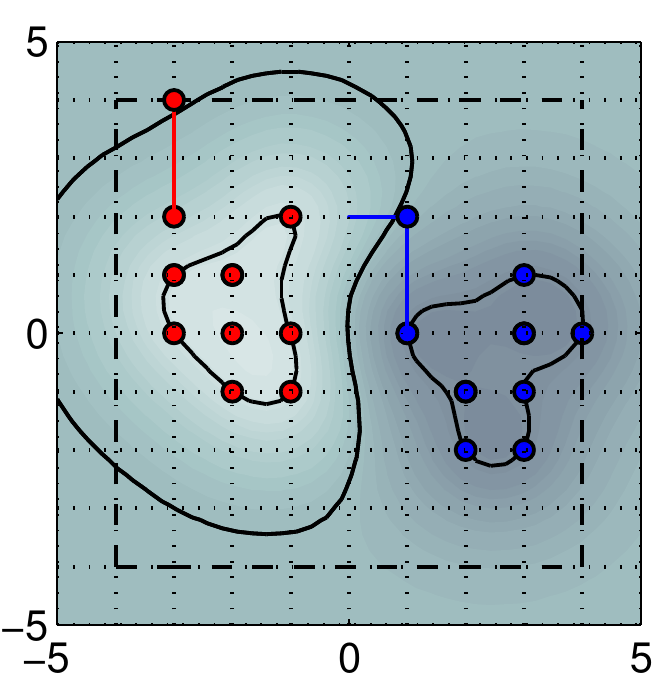} \hspace{0.25cm}
\includegraphics[height=0.2\textwidth]{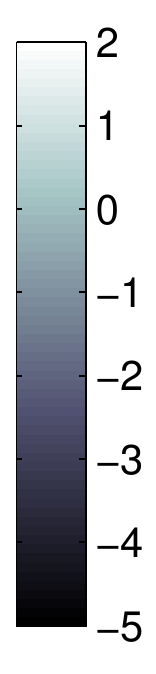}
\vspace{-5pt}
\caption{Examples of descent paths obtained by Algorithm~\ref{alg:classifier-security} to find optimal evasion points against SVM-based classifiers on a bi-dimensional dataset, for different choices of distance function, classification algorithm, and feature representation. Red and blue points represent the malicious and legitimate samples, respectively.
The initial malicious sample $\vct x$ to be modified in this example is the red point at $\vct x=[-3, 2]$.
Solid black lines denote the SVM's decision boundary (\ie, $g(\vct x)=0$) and margin conditions (\ie, $g(\vct x)=\pm 1$).
The color map represents the value of the discriminant function $g(\vct x)$ at each point.
Dashed black lines denote the boundaries of a box constraint.
Red and blue lines denote the descent paths when the initialization point is equal to the initial malicious point ($\vct x^{(0)}=\vct x$) and to the closest point classified as legitimate, respectively.
In the first and the second plot, evasion points are found by minimizing the $\ell_{2}$-norm $||\vct x^{\prime}-\vct x||_{2}^{2}$, respectively against a linear SVM with regularization parameter $C=1$, and against a nonlinear SVM with $C=1$ using the RBF kernel with $\gamma=0.5$, on a continuous feature space.
In the third and the fourth plot, the $\ell_{1}$-norm $||\vct x^{\prime}-\vct x||_{1}$ is minimized against the same nonlinear SVM, respectively on a continuous and on a discrete feature space, where feasible points lie at the grid intersections.
The problem may exhibit multiple local minima, or have a unique solution (see, \eg, the first plot).
Further, depending on the shape of the decision boundary, Algorithm~\ref{alg:classifier-security} may not find an evasion point when initialized with $\vct x^{(0)}=\vct x$ (see, \eg, the third plot). Accordingly, the closest point to $\vct x$ that evades detection should be eventually retained.}
\label{fig:2d-example}
\end{center}
\vspace{-12pt}
\end{figure*}

We explain here how to solve Problem~\eqref{eq:optimal-evasion}-\eqref{eq:optimal-evasion-2} to estimate the classifier security term $S(\vctg \theta)$ in the objective function of Eq.~\eqref{eq:wafs-criterion-1}. Problem~\eqref{eq:optimal-evasion}-\eqref{eq:optimal-evasion-2} essentially amounts to finding the closest sample $\vct x^{\prime}$ to $\vct x$ that evades detection, according to a given distance function $c(\vct x^{\prime}, \vct x)$.
In general, the problem may be solved using a black-box search approach (\eg, a genetic algorithm) that queries the classification function with different candidate samples to find the best evasion point.
This approach may be however too computationally demanding, as it does not exploit specific knowledge of the objective function and of the targeted classifier.
To develop more efficient algorithms, one should indeed focus on specific choices of the objective (\ie, the distance function), and of the constraints (\ie, the kind of classifier and feature representation).

As for the distance function, most of the work in adversarial learning has considered the $\ell_{1}$- and the $\ell_{2}$-norm, depending on the feature space and kind of attack~\cite{biggio13-ecml,nelson12-jmlr,lowd05,dalvi04}; \eg, if it is more convenient for an attacker to significantly manipulate few features than slightly manipulate the majority of them, the $\ell_{1}$-norm should be adopted, as it promotes \emph{sparsity}; otherwise, the $\ell_{2}$-norm would be a better choice.

The classification function may be linear (\eg, linear SVMs, perceptrons) or non-linear (\eg, SVMs with the radial basis function (RBF) or the polynomial kernel, neural networks). Further, it may be non-differentiable (\eg, decision trees).
Previous work has addressed the evasion of linear~\cite{lowd05} and convex-inducing classifiers, \ie, classifiers that partition the feature space into two sets, one of which is convex~\cite{nelson12-jmlr}.
It has also been recently shown that non-linear classifiers with differentiable discriminant functions can be evaded through a straightforward gradient descent-based attack~\cite{biggio13-ecml}. We are not aware of any work related to the evasion of non-linear classifiers with non-differentiable functions. Although this may be addressed through heuristic search approaches, as mentioned at the beginning of this section, we leave a more detailed investigation of this aspect to future work.
In this work we therefore consider classifiers whose discriminant function $g(\vct x)$ is not necessarily linear, but differentiable. These include, for instance, SVMs with differentiable kernels (\eg, linear, RBF, polynomial) and neural networks, which have been widely used in security-related applications~\cite{biggio14-tkde,biggio13-ecml}.

\begin{algorithm}[t]
\caption{Evasion Attack}
\label{alg:classifier-security}
\begin{algorithmic}[1]
\Require $\vct x$: the malicious sample;  $\vct x^{(0)}$: the initial location of the attack sample;
$t$: the gradient step size;  $\epsilon$: a small positive constant;
$m$: the maximum number of iterations.
\Ensure $\vct x'$: the closest evasion point to $\vct x$ found.
\State $i \gets 0$
\Repeat
    \State $i \gets i+1$
    \If{$g(\vct x^{(i)}) \geq 0$}{\; take a step towards the boundary}
	\State $\vct x^{(i)} \gets \vct x^{(i-1)} - t\nabla g(\vct x^{(i-1)}) $
    \Else{\; take a step to reduce the objective function}
    	\State $\vct x^{(i)} \gets \vct x^{(i-1)} - t\nabla c(\vct x^{(i-1)},\vct x) $
    \EndIf
    \If{$\vct x^{(i)}$ violates other constraints (\eg, box)}
    	\State Project $\vct x^{(i)}$ onto the feasible domain
    \EndIf
\Until{$c(\vct x^{(i)},\vct x) - c(\vct x^{(i-1)}, \vct x) < \epsilon$ or $i \geq m$}
\State \Return $\vct x^{\prime}= \vct x^{(i)}$
\end{algorithmic}
\end{algorithm}

Additional constraints to Problem~\eqref{eq:optimal-evasion}-\eqref{eq:optimal-evasion-2} may be considered, depending on the specific feature representation; \eg, if features are real-valued and normalized in $[0,1]^{d}$, one may consider a box constraint on $\vct x^{\prime}$, given as $0 \leq  x_{j}^{\prime} \leq 1$, for $j=1, \ldots, d$.
Further, features may take on discrete values, making our problem harder to solve, as discussed in Sect.~\ref{sect:discrete-features}.
In the following, we assume that feature values are continuous on a potentially compact space, such as $[0,1]^{d}$.

In the easiest cases, a solution can be found analytically; \eg, if one aims to minimize $c(\vct x^{\prime}, \vct x)=||\vct x^{\prime}-\vct x||^{2}_{2}$ against a linear classifier $g(\vct x)=\vct w^{\T} \vct x + b$ (being $\vct w \in \mathbb R^{d}$ and $b \in \mathbb R$ the feature weights and bias), it is easy to verify that the optimal evasion point is $\vct x^{\prime} = \vct x - g(\vct x) \frac{\vct w}{||\vct w||^{2}_{2}}$ (\cf{} Fig.~\ref{fig:2d-example}, leftmost plot).

If the discriminant function $g(\vct x)$ is non-linear, the optimization problem becomes non-linear as well, and it may thus exhibit local minima.
Nevertheless, a local minimum can be found by minimizing the objective function $c(\vct x^{\prime}, \vct x)$ through gradient descent.
Our idea is to take gradient steps that aim to reduce the distance of $\vct x^{\prime}$ from $\vct x$, while projecting the current point onto the feasible domain as soon as the constraint $g(\vct x^{\prime}) < 0$ is violated.
In fact, following the intuition in~\cite{biggio13-ecml}, the attack point can be projected onto the non-linear, feasible domain $g(\vct x^{\prime}) < 0$ by minimizing $g(\vct x^{\prime})$ itself through gradient descent. The detailed procedure is given as Algorithm~\ref{alg:classifier-security}.  Notice however that this projection may not always be successful (see Fig.~\ref{fig:2d-example}, third plot from the left). It may indeed happen that the attack point $\vct x^{\prime}$ reaches a flat region where the gradient of $g(\vct x^{\prime})$ is null, while the point is still classified as malicious ($g(\vct x^{\prime}) \geq 0$).
To overcome this limitation, we initialize the attack point to different locations before running the gradient descent (instead of mimicking the feature values of samples classified as legitimate, as done in~\cite{biggio13-ecml}). In particular, we consider two initializations: one in which the attack point $\vct x^{\prime}$ is set equal to $\vct x$ (red descent paths in Fig.~\ref{fig:2d-example}), and the other one in which $\vct x^{\prime}$ is set equal to the closest sample classified as legitimate (blue descent paths in Fig.~\ref{fig:2d-example}).
The rationale is the following. In the former case, we start from a point which is classified as malicious, and see whether we can reach a reasonably close evasion point by following the gradient of $g(\vct x^{\prime})$. In the latter case, we start from the closest point classified as legitimate, and try to get closer to $\vct x$ while avoiding violations of the constraint $g(\vct x^{\prime}) < 0$.
The closest point to $\vct x$ that evades detection is eventually retained. As shown in Fig.~\ref{fig:2d-example}, this should reasonably allow us to find at least a good local minimum for our problem.
Finally, note that the proposed algorithm quickly converges to the optimal evasion point when $g(\vct x^{\prime})$ is linear too, from any of the two proposed initializations, as shown in the leftmost plot of Fig.~\ref{fig:2d-example}.

\subsubsection{Gradients}
\label{sect:gradients}

The gradients required to evaluate classifier security using Algorithm~\ref{alg:classifier-security} are given below, for some distance and discriminant functions. Subgradients can be considered for non-differentiable functions, such as the $\ell_{1}$-norm.

\smallskip
\noindent \textbf{Distance functions.} As discussed in Sect.~\ref{sect:background}, typical choices for the distance function $c(\vct x^{\prime}, \vct x)$ in adversarial learning are the $\ell_{2}$- and the $\ell_{1}$-norm. Their gradients with respect to $\vct x^{\prime}$ can be respectively computed as $\nabla c(\vct x^{\prime}, \vct x) = 2 (\vct x^{\prime} - \vct x)$, and $\nabla c(\vct x^{\prime}, \vct x) = {\rm sign} (\vct x^{\prime} - \vct x)$, where ${\rm sign}(\vct v)$ returns a vector whose $i^{\rm th}$ element is $0$ if $v_{i}=0$, $1$ ($-1$) if $v_{i}>0$ ($v_{i}<0$).

\smallskip
\noindent \textbf{Linear classifiers.} For linear discriminant functions $g(\mathbf x) = \langle \mathbf w, \mathbf x \rangle + b$,  with feature weights $\mathbf w \in \mathbb{R}^{d}$ and bias $b \in \mathbb{R}$, the gradient is simply given as $\nabla g(\mathbf x) = \mathbf w$.

\smallskip
\noindent \textbf{Nonlinear SVMs.} For kernelized SVMs, the discriminant function is $g(\mathbf x) = \sum_{i=1}^{n} \alpha_i y_i k(\mathbf x,\mathbf x_{i}) + b$, where the parameters $\vctg \alpha$ and $b$ are learned during training, $k(\mathbf x,\mathbf x_{i})$ is the kernel function, and $\{\vct x_{i}, y_{i}\}_{i=1}^{n}$ are the training samples and their labels~\cite{vapnik95-book}. The gradient is $\nabla g(\mathbf x) = \sum_i \alpha_i y_i \nabla k(\mathbf x,\mathbf x_{i})$. In this case, the feasibility of our approach depends on the kernel derivative $\nabla k(\mathbf x,\mathbf x_{i})$, which is computable for many numeric kernels; \eg, for the RBF kernel $k(\mathbf x,\mathbf x_{i}) = \exp\{-\gamma \|\mathbf x-\mathbf x_{i}\|^2\}$, it is $\nabla k(\mathbf x,\mathbf x_{i}) = -2 \gamma \exp\{-\gamma \|\mathbf x-\mathbf x_{i}\|^2\}(\mathbf x-\mathbf x_{i})$. 

Although in this work we only consider kernelized SVMs as an example of nonlinear classification, our approach can be easily extended to any other nonlinear classifier with differentiable discriminant function $g(\vct x)$ (see, \eg, \cite{biggio13-ecml}, for the computation of $\nabla g(\vct x)$ for neural networks).

\subsubsection{Descent in discrete spaces}
\label{sect:discrete-features}

In discrete spaces, it is not possible to follow the gradient-descent direction exactly, as this may map the current sample to a set of non-admissible feature values.
In fact, descent in discrete spaces amounts to finding, at each step, a feasible neighbor of the current sample that maximally decreases the objective function.
This can be generally addressed using a search algorithm that queries the objective function at every point in a small neighborhood of the current sample, which may however require a large number of queries (exponential in the number of features).
For classifiers with a differentiable discriminant function, the number of queries can be reduced by perturbing only
a number of features which correspond to the maximum absolute values of the gradient, one at a time, in the correct direction, and eventually retaining the sample that maximally decreases the objective function.
This basically amounts to exploiting the available gradient as a search heuristic, and to selecting the feasible point that best aligns with the current gradient. An example of the descent paths explored in discrete spaces by our evasion attack algorithm is given in Fig.~\ref{fig:2d-example} (rightmost plot).


\section{Security Evaluation}
\label{sect:attacker-model}

To evaluate the effectiveness of the proposed adversarial feature selection method against traditional feature selection approaches, we follow the security evaluation procedure originally proposed in~\cite{biggio14-tkde,biggio13-ecml}.
The underlying idea is to simulate attacks that may be potentially incurred at test time, relying on specific assumptions on the adversary's model, in terms of his goal, knowledge of the targeted system, and capability of manipulating the input data.
In the \emph{evasion} setting, the attacker aims to evade detection by exploiting knowledge of the classification function to manipulate the malicious (test) samples.
The attacker's knowledge can be either \emph{perfect} or \emph{limited}.
In the former case, the classification algorithm is fully known to the attacker, who can then perform a worst-case attack,
similarly to Problem~\eqref{eq:optimal-evasion}-\eqref{eq:optimal-evasion-2}.
In the latter case, instead, knowledge of the \emph{true} discriminant function $g(\vct x)$ is not available. This is a more realistic case, as typically the attacker has neither access to the classifier internal parameters nor to the training data used to learn it.
Nevertheless, the function $g(\vct x)$ can be approximated by collecting \emph{surrogate} data (\ie, data ideally sampled from the same distribution followed by the training data used to learn the targeted classifier), and then learning a surrogate classifier on it. As shown in~\cite{biggio13-ecml}, this can lead to very good approximations of the targeted classifier for the sake of finding suitable evasion points.
Intuitively, solving Problem~\eqref{eq:optimal-evasion}-\eqref{eq:optimal-evasion-2} when exploiting an approximation $\hat g(\vct x)$ of the true discriminant function $g(\vct x)$ may not be a good choice, as looking for evasion points which are only barely misclassified as legitimate by the \emph{surrogate} classifier may lead the attacker to only rarely evade the \emph{true} classifier. For this reason, optimal evasion has been reformulated in~\cite{biggio13-ecml} as:
\begin{eqnarray}
\label{eq:sec-eval-1} \min_{\vct x^{\prime}} && \hat g(\vct x^{\prime}) \enspace , \, \\
\label{eq:sec-eval-2} {\rm s.t.} && c(\vct x^{\prime}, \vct x) \leq c_{\rm max} \enspace ,
\end{eqnarray}
where the constraint on $c(\vct x^{\prime}, \vct x)$ bounds the attacker's capability by setting an upper bound on the maximum amount of modifications $c_{\rm max}$ that can be made to the initial malicious sample $\vct x$.
In this case, the malicious sample is modified to be misclassified as legitimate with the highest possible confidence (\ie, minimum value of $\hat g(\vct x)$), under a maximum amount of modifications $c_{\rm max}$, which can be regarded as a parameter of the security evaluation procedure.
In fact, by repeating the security evaluation for increasing values of $c_{\rm max}$, one can show how gracefully the performance of the true classifier decreases against attacks of increasing \emph{strength}.
The more the performance gracefully decreases, the more secure the classifier is expected to be.
Examples of such curves will be shown in Sect.~\ref{sect:experiments}.
Finally, note that the aforementioned problem can be solved using an algorithm similar to Algorithm~\ref{alg:classifier-security}, in which the objective function and the constraint are exchanged.


\section{Application Examples}
\label{sect:experiments}

In this section we empirically validate the proposed approach on two application examples involving spam filtering and PDF malware detection.
In the former case, we compare the traditional \emph{forward} feature selection wrapping algorithm with the corresponding implementation of our approach, using a linear SVM as the classification algorithm.
In the latter case, instead, we consider traditional and adversarial \emph{backward} feature elimination approaches, and an SVM with the RBF kernel as the wrapped classifier.
We believe that these examples can be considered a representative set of cases to assess the empirical performance of the proposed method.

\vspace{-10pt}
\subsection{Spam Filtering}
\label{sect:exp:spam}


Spam filtering is one of the most common application examples considered in adversarial machine learning~\cite{biggio14-tkde,nelson09,huang11,rubinstein09,biggio-IJMLC10}.
In this task, the goal is often to design a \emph{linear} classifier that discriminates between legitimate and spam emails by analyzing their textual content, exploiting the so-called \emph{bag-of-words} feature representation, in which each binary feature denotes the presence (1) or absence (0) of a given word in an email~\cite{manning1999}.
Despite its simplicity, this kind of classifier has shown to be highly accurate, while also providing interpretable decisions. It has been therefore widely adopted in previous work~\cite{biggio14-tkde,nelson09,huang11,rubinstein09,biggio-IJMLC10,jorgensen08,lee2005spam,kolcz09}, and in several real anti-spam filters, like SpamAssassin and SpamBayes.\footnote{\url{http://spamassassin.apache.org/}, \url{http://spambayes.sourceforge.net/}}
Evasion attacks against these kinds of classifier consist of manipulating the content of spam emails through bad word obfuscations (\eg, misspelling spammy words like ``cheap'' as ``che4p'') and good word insertions (\ie, adding words which typically appear in legitimate emails and not in spam), which amounts to modifying the corresponding feature values from 1 to 0 and vice versa~\cite{kolcz09,biggio-IJMLC10,lowd05,biggio14-tkde}.

\textbf{Experimental setup.} For these experiments, we considered the benchmark TREC 2007 email corpus, which consists of 25,220 legitimate and 50,199 real spam emails~\cite{trec07}.
We represented each email as a feature vector using the tokenization method of SpamAssassin, which exploits the aforementioned bag of-words feature model. To this end, we first extracted the dictionary of words (\ie, features) from the first 5,000 emails (in chronological order). Then, to keep the computational complexity manageable, we reduced the feature set from more than 20,000 to 500  features, without significant loss in classification accuracy, using a supervised feature selection approach based on the information gain criterion~\cite{Fabrizio01}.\footnote{Note that a similar procedure has also been carried out in \cite{biggio14-tkde}, where it is also quantitatively shown that classification performance is not affected.}
The linear SVM was considered as the classification algorithm.
As for the performance measure $G(\vctg \theta)$, we used classification accuracy. Classifier security $S(\vctg \theta)$ was evaluated as discussed in Sect.~\ref{sect:adv-feat-sel}, using the $\ell_{1}$-norm as the distance function $c(\vct x^{\prime},\vct x)$, and Algorithm~\ref{alg:classifier-security} for discrete spaces (Sect.~\ref{sect:discrete-features}). This choice of distance function amounts to counting the minimum number of words to be modified in each spam to evade detection.

We run a preliminary security evaluation to tune the trade-off parameter $\lambda$ of our method, aiming to maximize the average true positive (TP) rate (\ie, the fraction of correctly classified malicious samples) at the 1\% false positive (FP) rate operating point (\ie, when 1\% of the legitimate samples are misclassified), for $c_{\rm max} \in [0,20]$ (see Sect.~\ref{sect:attacker-model}). It is indeed common to evaluate system performance at low FP rates in security-related tasks~\cite{kolcz09,biggio-IJMLC10,biggio14-tkde}. This also allows us to compare different classifiers against evasion attacks in a fair way, as classifiers with higher FP rates may be easier to evade~\cite{biggio14-tkde,biggio13-ecml}.
If $\lambda$ is too large, the selected features show poor generalization capability in the absence of attack (\ie, when $c_{\rm max}=0$), which also leads to a too low TP rate under attack (\ie, when $c_{\rm max}>0$). Conversely, if $\lambda$ is too small, classifier performance under attack may quickly decrease. Hence, to effectively tune $\lambda$, one should quantitatively investigate this trade-off on the available data. We assume that a maximum decrease of the TP rate in the absence of attack of $1\%$ is tolerable, at the given FP rate. Then, the highest value of $\lambda$ under this constraint can be selected, to maximize the TP rate under attack. Based on these observations, we run a 5-fold cross validation on the training set with values of $\lambda \in \{0.1, 0.5, 0.9\}$, and selected $\lambda=0.5$.

Each experiment was repeated five times, each time by randomly selecting 2,500 samples for each class from the remaining emails in the TREC corpus.
In each run, the dataset of 5,000 samples was randomly split into a training and a test set of 2,500 samples each.
Then, subsets consisting of 1 to 499 features were selected according to the traditional and adversarial forward selection methods, through a 5-fold cross validation procedure on the training set, to respectively maximize the value of $G(\vctg \theta)$ and $G(\vctg \theta) + \lambda S(\vctg \theta)$ (Algorithm~\ref{alg:wafs}) estimated on such data.
The SVM regularization parameter $C \in \{2^{-10}, 2^{-9}, ..., 2^{10}\}$ was also selected during this process using an additional inner 5-fold cross-validation to maximize classification accuracy, usually yielding $C=1$.

Security evaluation was then carried out on the test data. Similarly to the procedure used to tune $\lambda$, we manipulated  each malicious sample according to Problem~\eqref{eq:sec-eval-1}-\eqref{eq:sec-eval-2} for $c_{\rm max} \in [0,20]$, assuming perfect (PK) and limited (LK) knowledge of the \emph{true} discriminant function $g(\vct x)$.
As discussed in Sect.~\ref{sect:attacker-model}, in the LK case the attacker constructs evasion points by attacking a surrogate classifier $\hat g(\vct x)$. These points are then used to attack the \emph{true} classifier $g(\vct x)$ and evaluate the performance. To provide a realistic evaluation, as in \cite{biggio13-ecml}, we learn the surrogate classifier using a smaller training set, consisting of only 500 samples.
For each value of $c_{\rm max}$ we then computed TP at 1\% FP for the true classifier under the PK and LK attack scenarios.



\begin{figure*}[t]
\centering
\includegraphics[width=0.227\textwidth]{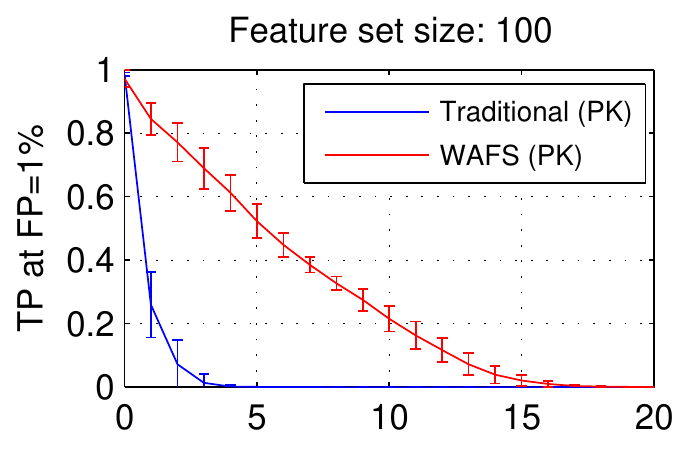} \hspace{0.22cm}
\includegraphics[width=0.227\textwidth]{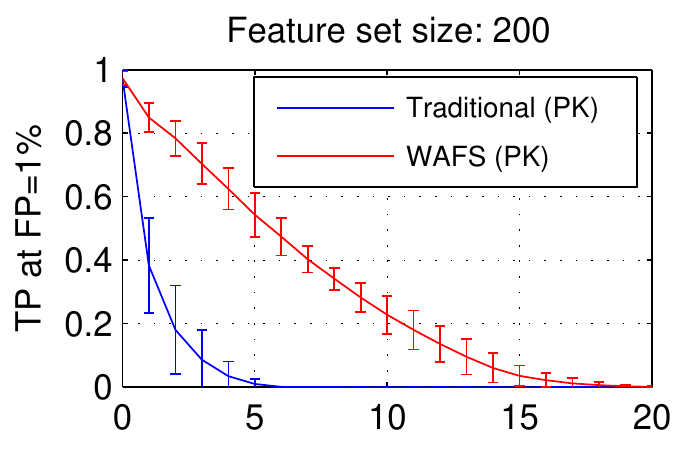} \hspace{0.22cm}
\includegraphics[width=0.227\textwidth]{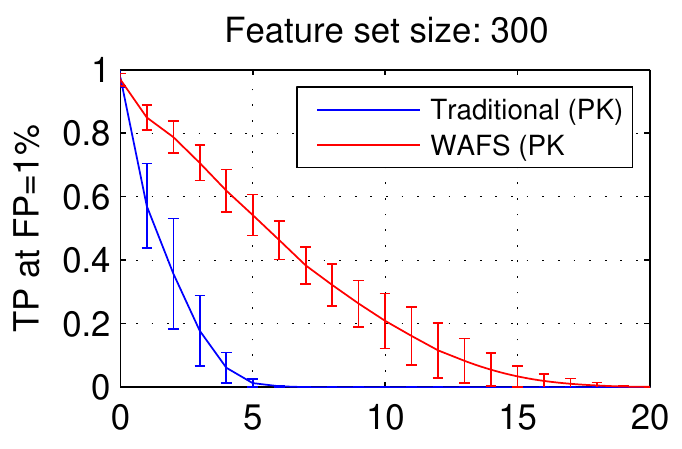} \hspace{0.22cm}
\includegraphics[width=0.227\textwidth]{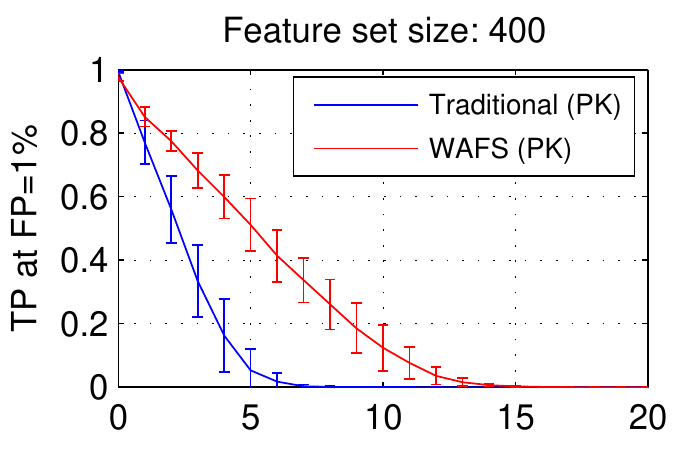}
\includegraphics[width=0.227\textwidth]{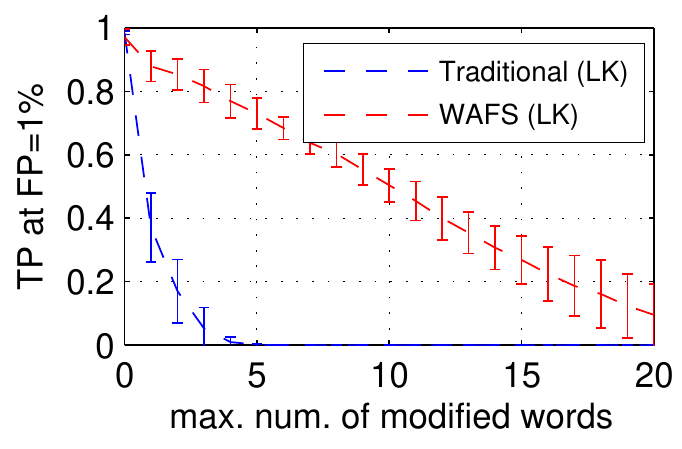} \hspace{0.25cm}
\includegraphics[width=0.227\textwidth]{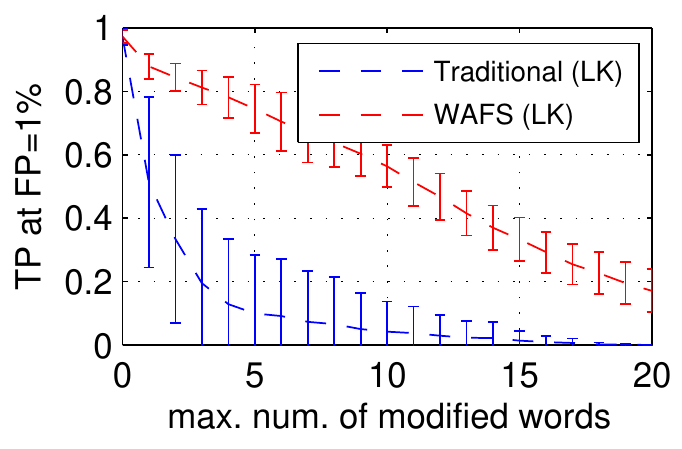} \hspace{0.25cm}
\includegraphics[width=0.227\textwidth]{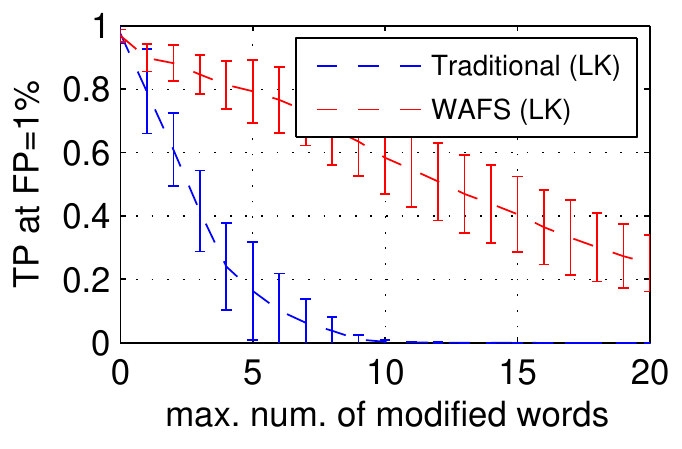} \hspace{0.25cm}
\includegraphics[width=0.227\textwidth]{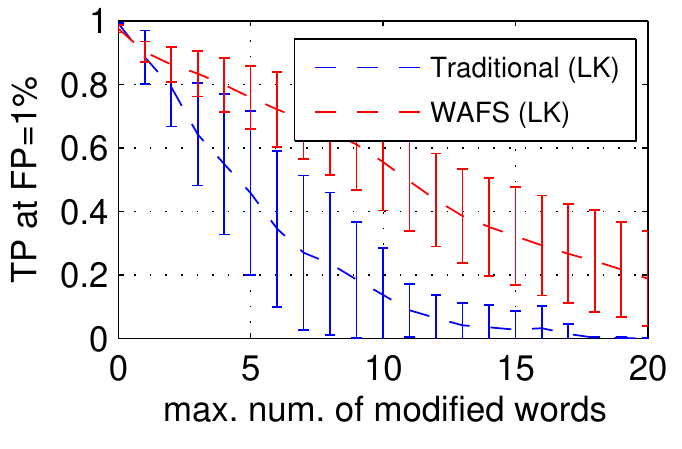}
\vspace{-5pt}
\caption{Security evaluation curves for the spam data, showing the average and standard deviation of the TP rate at 1\% FP rate, for feature subset sizes of 100, 200, 300, and 400 (in different columns), and for the PK (top row) and LK (bottom row) attack scenarios.}
\label{fig:exp-sec-eval-spam}
\vspace{-10pt}
\end{figure*}

\textbf{Experimental results.} The average value (and standard deviation) of TP at 1\% FP for the security evaluation procedure described above are reported in Fig.~\ref{fig:exp-sec-eval-spam}, for feature subset sizes of 100, 200, 300, and 400, and for the PK and LK attack scenarios.
In the absence of attack (\ie, when $c_{\rm max}=0$), the two methods exhibited similar performances. Although the traditional method performed occasionally  better than WAFS, the difference turned out not to be 95\% statistically significant according to the Student's t-test. Under attack (\ie, when $c_{\rm max}>0$), instead, WAFS always significantly outperformed the traditional method, for both the PK and the LK attack scenarios.
As the performance of the traditional method decreased less gracefully as $c_{\rm max}$ increased, we can conclude that WAFS leads to learning more secure classifiers. This is also confirmed by the LK attack scenario. In this case, even if only a surrogate classifier is available to the attacker, manipulating up to 20 words in a spam email may allow one to evade the true classifier almost surely (provided that the selected features are known to the attacker).

\begin{figure}[!htbp]
\centering
\includegraphics[width=0.45\textwidth]{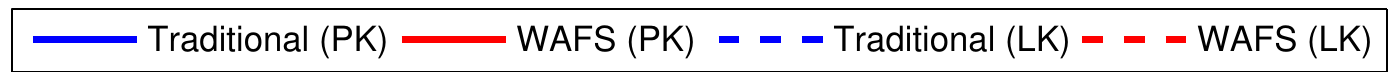}\\
\includegraphics[width=0.22\textwidth]{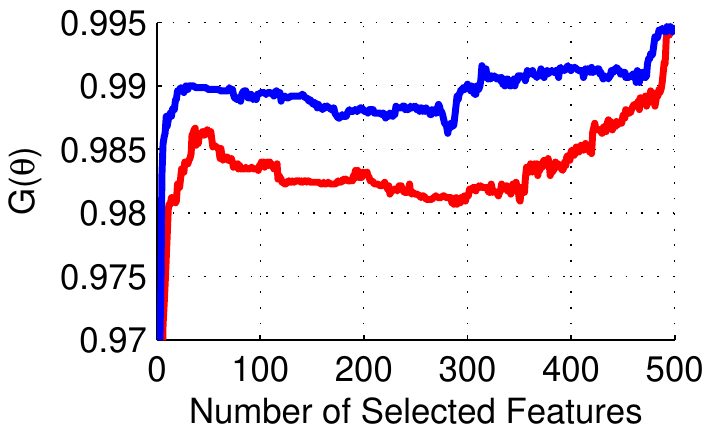}
\includegraphics[width=0.22\textwidth]{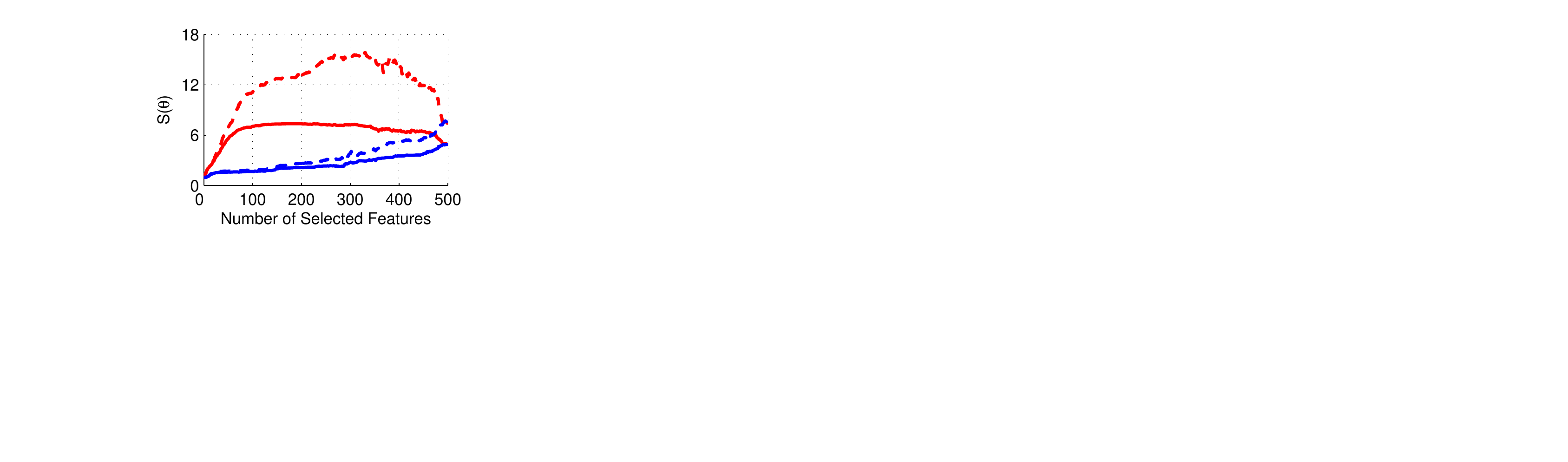}
\caption{Average classification accuracy $G(\vctg \theta)$ in the absence of attack (left plot), and classifier security $S(\vctg \theta)$ under attack (\ie, average minimum number of modified words in each spam to evade detection)  for the PK and LK attack scenarios (right plot), against different feature subset sizes, for the spam data.}
\label{fig:exp-selection-spam}
\vspace{-12pt}
\end{figure}

In Fig.~\ref{fig:exp-selection-spam}, we also report the values of $G(\vctg \theta)$ (\ie,  the classification accuracy in the absence of attack) and $S(\vctg \theta)$ (\ie, the average minimum number of features to be modified in a malicious sample to evade detection). Note how the proposed adversarial feature selection method systematically required the attacker to modify a higher number of features (\ie, words) to evade detection, for all considered feature subset sizes, without significantly affecting the accuracy in the absence of attack $G(\vctg \theta)$. This clearly confirms our previous results, showing that maximizing the security term $S(\vctg \theta)$ during feature selection helps improving the detection rate under attack, and thus classifier security. Although this comes at the cost of an increased computational complexity (since it requires simulating attacks against the targeted classifier), feature selection is often carried out offline, and thus the additional running time required by our method may not be critical.

Finally, we evaluated the correlation of the \emph{hardness of evasion}, \ie, the classifier security measure $S(\vctg \theta)$ used in this work, with the \emph{weight evenness}~\cite{kolcz09,biggio-IJMLC10}, \ie, another possible measure for assessing the security of linear classifiers (see Sect.~\ref{sect:background-classifier-security}). The goal is to verify whether the weight evenness can be adopted to compute the security term $S(\vctg \theta)$ in our approach, as it can be computed more efficiently than the hardness of evasion, \ie, without simulating any attack against the trained classifier.
To this end, we trained 200 linear classifiers using 200 distinct samples each, and evaluated the correlation between the two considered measures.
Surprisingly, our experiment showed that the two measures were not significantly correlated (the Pearson's correlation coefficient was almost zero), contradicting the intuition in previous studies~\cite{kolcz09,biggio-IJMLC10}. One possible reason is that the weight evenness does not exploit any information on the data distribution besides the classifier's feature weights. Therefore, it may not be properly suited to characterize classifier security.

\vspace{-8pt}
\subsection{Malware Detection in PDF Files}
\label{sect:exp-pdf}

Here, we consider another realistic application example related to the detection of malware (\ie, \emph{mal}icious soft\emph{ware}) in PDF files.
These files are characterized by a hierarchy of interconnected objects,
each consisting of a \emph{keyword}, denoting its type, and a \emph{data stream}, representing its content; \eg, the keyword \texttt{/PageLayout} characterizes an object describing how the correponding page is formatted.
This flexible, high-level structure allows for embedding of different kinds of content, such as \texttt{JavaScript}, \texttt{Flash}, and even binary programs, which in turn makes PDF files particularly attractive as vectors for disseminating malware.
Recent work has exploited machine learning as a tool
for detecting malicious PDF files based on their logical
structure; in particular, on the set of embedded keywords~\cite{maiorca13-asiaccs,srndic13-ndss}.
In this application example, we use the same feature representation
exploited in~\cite{maiorca13-asiaccs,biggio13-ecml},
in which each feature represents the number of occurrences of a given keyword in a PDF file.
Conversely to the spam filtering example, feature values in this case can not be modified in an unconstrained manner to perform an evasion attack.
In fact, it is not to trivially possible to remove an embedded object (and
the associated keywords) from a PDF without corrupting its
structure. Nevertheless, it is quite easy to add new objects (\ie, keywords)
through the PDF versioning mechanism (see \cite{maiorca13-asiaccs,biggio13-ecml} and references therein).
In our case, this can be easily accounted for by setting $\vct x \leq \vct x^{\prime}$ as an additional constraint to Problem~\eqref{eq:optimal-evasion}-\eqref{eq:optimal-evasion-2} and Problem~\eqref{eq:sec-eval-1}-\eqref{eq:sec-eval-2}, respectively for the purpose of finding the optimal evasion points, and running the security evaluation procedure.\footnote{With the notation $\vct x \leq \vct x^{\prime}$, we mean that $x_{j} \leq x_{j}^{\prime}$, for $j=1,\ldots,d$.}

\begin{figure*}[t]
\centering
\includegraphics[width=0.227\textwidth]{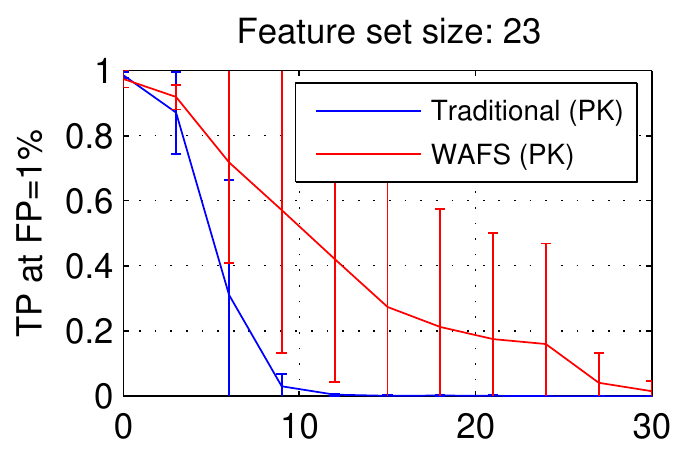}  \hspace{0.22cm}
\includegraphics[width=0.227\textwidth]{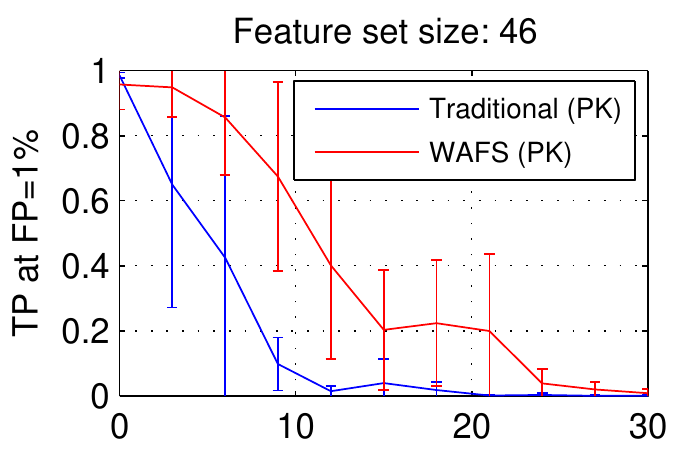}  \hspace{0.22cm}
\includegraphics[width=0.227\textwidth]{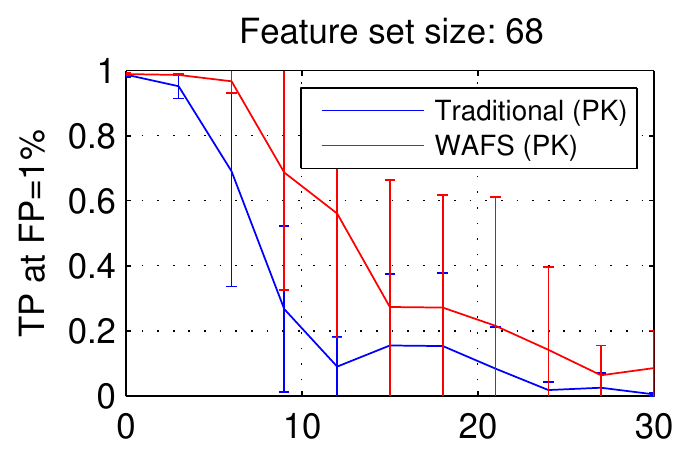}  \hspace{0.22cm}
\includegraphics[width=0.227\textwidth]{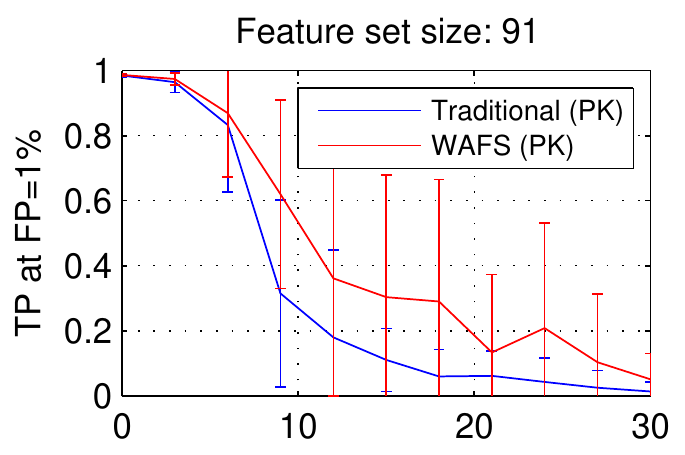}
\includegraphics[width=0.227\textwidth]{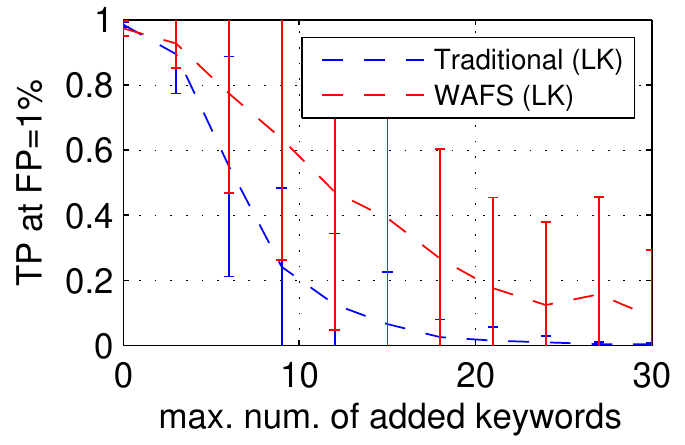}  \hspace{0.25cm}
\includegraphics[width=0.227\textwidth]{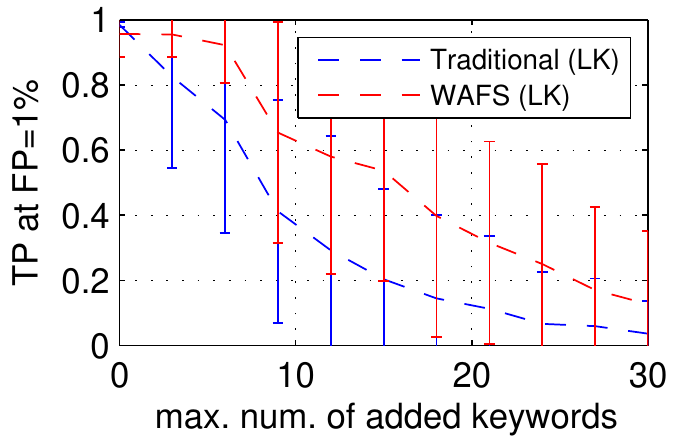}  \hspace{0.25cm}
\includegraphics[width=0.227\textwidth]{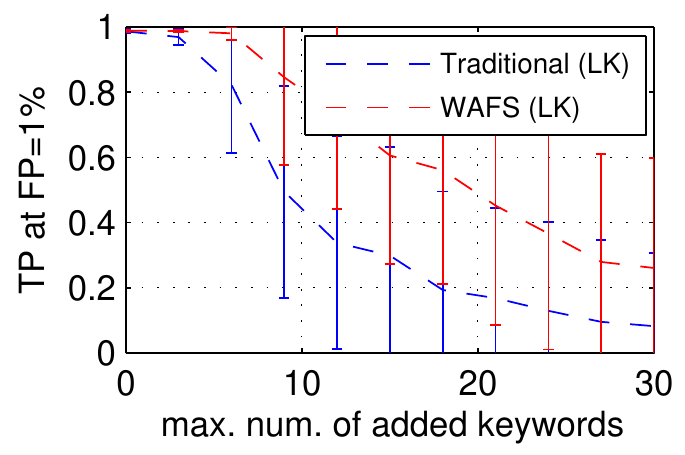}  \hspace{0.25cm}
\includegraphics[width=0.227\textwidth]{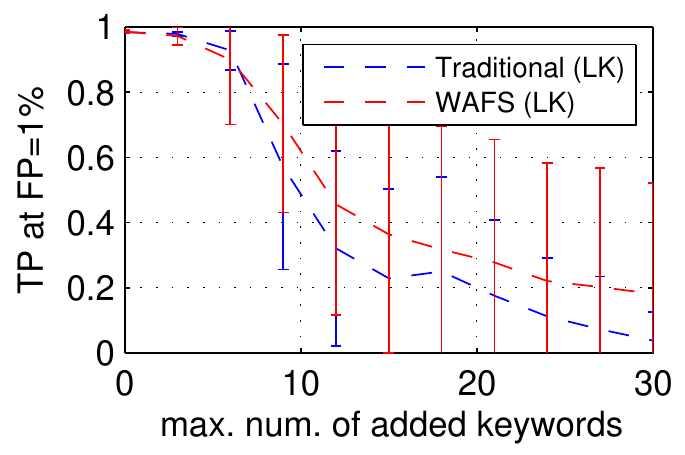}
\vspace{-5pt}
\caption{Security evaluation curves for the PDF malware data, showing the average and standard deviation of the TP rate at 1\% FP rate, for feature subset sizes of 23, 46, 68, and 91 (in different columns), and for the PK (top row) and LK (bottom row) attack scenarios.}
\label{fig:exp-sec-eval-pdf}
\vspace{-10pt}
\end{figure*}

\textbf{Experimental setup.} In these experiments, we considered the PDF malware dataset used in \cite{maiorca13-asiaccs,biggio13-ecml}, including 5591 legitimate and 5993 malicious PDFs.
As mentioned above, features have integer values, each representing the occurrence of a given keyword in a PDF.
In total, 114 distinct keywords were found from the first 1,000 samples (in chronological order). They were used as our set of features. As in \cite{biggio13-ecml}, we limited the influence of outlying observations by setting the maximum value of each feature to 100.
We then normalized each feature by simply dividing its value by 100.
The SVM with the RBF kernel was used as the classification algorithm.
The same performance $G(\vctg \theta)$ and classifier security $S(\vctg \theta)$ measures defined in the previous section were considered here.
Notably, using the $\ell_{1}$-norm to evaluate $S(\vctg \theta)$ in this case amounts to counting the number of added keywords to a PDF (divided by 100).
It is clear that the feature space is discrete in this case as well, since the admissible values are $\vct x \in \{0, 1/100, 2/100, \ldots, 1\}^{d}$. We therefore exploited again Algorithm~\ref{alg:classifier-security} on discrete spaces (Sect.~\ref{sect:discrete-features}) for the sake of estimating $S(\vctg \theta)$ and running security evaluation.

Following the same experimental setup used for the spam filtering task, we set $\lambda=0.9$.
Experiments were repeated ten times, each time randomly drawing 1,000 samples from the remaining data.
In each run, these samples were split into a training and a test set of equal sizes.
Then, feature subsets of sizes from 1 to 113 were selected using traditional and adversarial backward feature elimination, performing a 5-fold cross-validation on the training data.
The SVM regularization parameter $C \in \{2^{-10}, 2^{-9}, ..., 2^{10}\}$ and the  kernel parameter $\gamma \in \{2^{-3}, 2^{-2}, ..., 2^{3}\}$, were set during this process through an inner 5-fold cross-validation, maximizing the classification accuracy. This typically yielded $C=256$ and $\gamma=0.5$.
Security evaluation was performed as for the previous experiments on spam filtering, considering $c_{\rm max} \in [0,0.5]$, which amounts to adding a maximum of 50 keywords to each malicious PDF.

\begin{figure}[t]
\centering
\includegraphics[width=0.45\textwidth]{figs/Legend.pdf}\\
\includegraphics[width=0.22\textwidth]{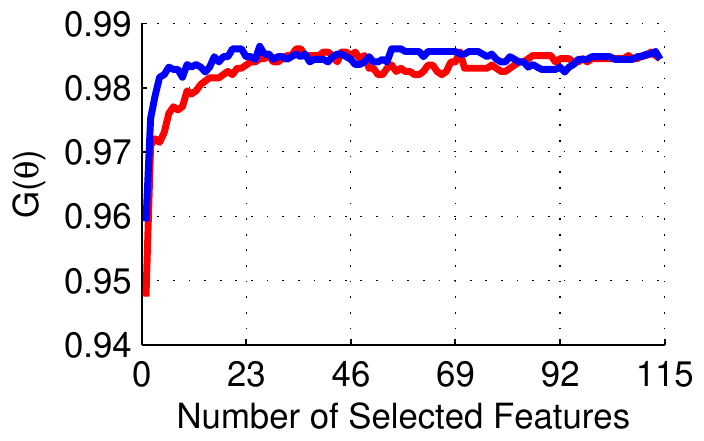}
\includegraphics[width=0.22\textwidth]{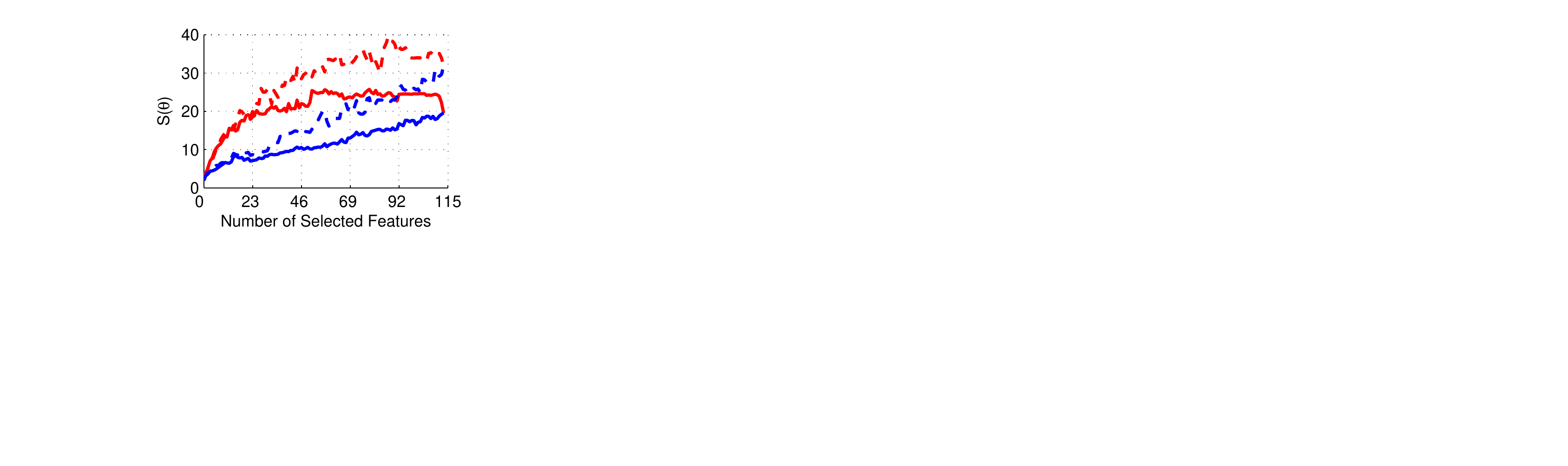}
\caption{Average classification accuracy $G(\vctg \theta)$ in the absence of attack (left plot), and classifier security $S(\vctg \theta)$ under attack (\ie, average minimum number of added keywords to each malicious PDF to evade detection) for PK and LK attacks (right plot), against different feature subset sizes, for the PDF data.}
\label{fig:exp-selection-pdf}
\vspace{-12pt}
\end{figure}


\textbf{Experimental results.} In Fig.~\ref{fig:exp-sec-eval-pdf}, we report the average value (and standard deviation) of TP at 1\% FP for the SVM with the RBF kernel trained on feature subsets of 23, 46, 68 and 91 features (\ie, 20\%, 40\%, 60\% and 80\% of the total number of features), selected by the traditional backward elimination wrapping algorithm and by the corresponding WAFS implementation, under the PK and LK attack scenarios. Similar observations to those reported for the spam filtering task can be made here; in particular, WAFS outperformed traditional feature selection in terms of TP values under attack, without exhibiting a significant performance loss in the absence of attack.
Moreover, in the LK case, a higher number of added keywords was required, as expected, to reach a comparable detection rate to that reported for the PK attack scenario.

It is worth noting here that the variance of the TP rates reported in Fig.~\ref{fig:exp-sec-eval-pdf} turned out to be significantly higher than in the previous experiments. This fluctuation might be due to the use of smaller training sets, and to the higher variability induced by the use of a nonlinear decision boundary.
Consequently, only few cases were 95\% statistically significant based on the Student's t-test in the PK scenario. In the LK scenario, as the difference between the average values of the two methods was larger, WAFS was 95\% significantly better than the traditional backward elimination algorithm in all cases for feature subsets of 23, 46 and 68 features, except for $c_{\rm max} \in [0,0.03]$.
Although some results were not 95\% statistically significant due to the high variability of our results, we can nevertheless conclude that WAFS  was able to outperform the traditional backward elimination approach in most of the cases.

The classification accuracy in the absence of attack $G(\vctg \theta)$, and the average minimum number of keywords added to a malicious PDF to evade detection (\ie, $S(\vctg \theta) \times 100$) for the SVM with the RBF kernel trained on features selected by the traditional and the adversarial feature selection methods are shown in Fig.~\ref{fig:exp-selection-pdf}. Similarly to the results reported in the spam filtering example, the $S(\vctg \theta)$ values for WAFS are significantly higher than those exhibited by the traditional feature selection method in both the PK and the LK scenarios, although the SVM's classification accuracy $G(\vctg \theta)$ is not significantly affected. It should however be noted that the additional computational complexity required to compute $S(\vctg \theta)$ for a nonlinear classifier is higher than that required by a linear classifier due to the intrinsic complexity of exploring a nonlinear decision boundary.
Finally, it is worth pointing out that WAFS was able to improve classifier security in this case by mainly selecting features that exhibited, on average, higher values for the malicious class. In fact, due to the constraint $\vct x \leq \vct x^{\prime}$, it becomes harder for an attacker to mimic characteristic feature values of the legitimate class in this case, yielding eventually a lower probability of evading detection. This may be an interesting research direction to explore, in order to devise surrogate measures of classifier security suitable for implementing adversarial feature selection as a more computationally efficient filter method.


%

\vspace{-5pt}
\section{Conclusions and Future Work}
\label{sect:conclusions}

Feature selection may be considered a crucial step in security-related applications, such as spam and malware detection, when small subsets of features have to be selected to reduce computational complexity, or to improve classification performance by tackling the course of dimensionality~\cite{GuyonIsabelle2003,brown12}.
However, since traditional feature selection methods implicitly assume that training and test samples follow the same underlying data distribution,
their performance may be significantly affected under adversarial attacks that violate this assumption. Even worse, performing feature selection in adversarial settings may allow an attacker to evade the classifier at test time with a lower number of modifications to the malicious samples~\cite{biggio-IJMLC10,biggio14-tkde,bo14-nips,wang14-icdm}.
To our knowledge, besides the above studies,
the issue of selecting feature sets suitable for adversarial settings has neither been experimentally nor theoretically investigated more in depth.

In this paper, we proposed an \emph{adversarial} feature selection method that optimizes not only the generalization capability of the wrapped classifier, but also its security against evasion attacks at test time.
To this end, we extended a previous definition of classifier security, which was suited to linear classifiers trained on binary features, to the case of nonlinear classification algorithms trained on either continuous or discrete feature spaces.
We validated the soundness of our approach on realistic application examples involving spam and PDF malware detection. We showed that the proposed approach can outperform traditional approaches in terms of classifier security, without significantly affecting the classifier performance in the absence of attacks.
We also empirically showed that the proposed measure of classifier security provides a better characterization of this aspect than other previously-proposed measures aimed at evaluating the security of linear classifiers.
However, our method demands for an increased computational complexity, with respect to traditional wrapping algorithms,
as it requires simulating evasion attacks against the wrapped classifier at each iteration of the feature selection process.
Although this may not be a critical aspect, as feature selection is often carried out offline, making our approach more efficient remains an open issue to be investigated in future work.

A possible solution to overcome this limitation, and exploit our method in the context of more efficient feature selection approaches like \emph{filter} methods, may be to devise suitable surrogate measures of classifier security that can reliably approximate this value without simulating attacks against the trained classifier. Investigating the connections between security and stability of the feature selection process may be one fruitful research direction to this end, as  discussed in Sect.~\ref{sect:background}.
A more concrete example is however given at the end of Sect.~\ref{sect:exp-pdf}, based on the intuition of restricting the feasible space of sample manipulations available to the attacker. In practice, if the feature values of malicious samples can only be incremented, an adversarial feature selection procedure should prefer selecting features that exhibit lower values for samples in the legitimate class, making thus harder for an attacker to evade detection by mimicking such samples. This can be easily encoded by a measure that does not require training and attacking the corresponding classifier, and that can be thus exploited in the context of filter-based feature selection.

Another interesting extension of this work is related to the application of the proposed approach in the context of more complex feature mappings, \ie, feature spaces in which there is not a clear, direct relationship with the characteristics of each sample, and therefore it is not trivial to understand how to modify a malicious sample to find an optimal evasion point, \ie, to exhibit the desired feature values. This is however an application-specific issue, known in the adversarial machine learning literature as the inverse feature-mapping problem~\cite{huang11,biggio14-tkde}. In practice, the problem arises from the fact that optimal attacks are defined in feature space, and thus finding the corresponding  optimal \emph{real} samples requires reverse engineering the feature mapping.
From a pragmatic perspective, this can be overcome by first defining a suitable set of manipulations that can be made to the real malicious samples (\eg, many tools are available to obfuscate the content of malware samples, by manipulating their code~\cite{maiorca13-asiaccs}), and considering such manipulations as the only ones available to the attacker to find evasion points. Although this may lead us to find only suboptimal evasion points in feature space (as we restrict the attacker to work on a potentially smaller feasible set), we are guaranteed that the corresponding \emph{real} samples not only exist, but they are also easily determined.

To conclude, we believe that our work provides a first, concrete attempt towards understanding the potential vulnerabilities of feature selection methods in adversarial settings, and towards developing more secure feature selection schemes against adversarial attacks.

\section*{Acknowledgments}
The authors would like to thank Davide Maiorca for providing them the PDF malware dataset.
This work was supported in part by the National Natural Science Foundation of China under Grant 61003171, Grant 61272201, and Grant 61003172, and in part by Regione Autonoma della Sardegna under Grant CRP-18293, L. R. 7/2007, Bando 2009.

\vspace{-5 pt}

\vspace{-1.2 cm}
\begin{IEEEbiography}
[{\includegraphics[width=1in,height =1.25in,clip,keepaspectratio]{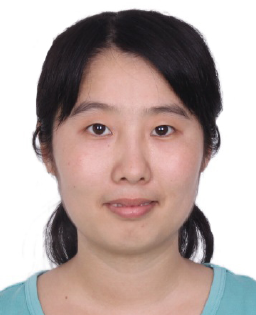}}]{Fei Zhang (S'11)} received her B.S. degree in Information and Computer Science from Minnan Normal University, Zhangzhou, China, in 2009. Now she is a Ph. D. student in South China University of Technology. Her current interested research is focused on machine learning, computer security and spam filtering. Miss Zhang is an IEEE student member.
\end{IEEEbiography}

\vspace{-1 cm}

\begin{IEEEbiography}
[{\includegraphics[width=1in,height =1.25in,clip,keepaspectratio]{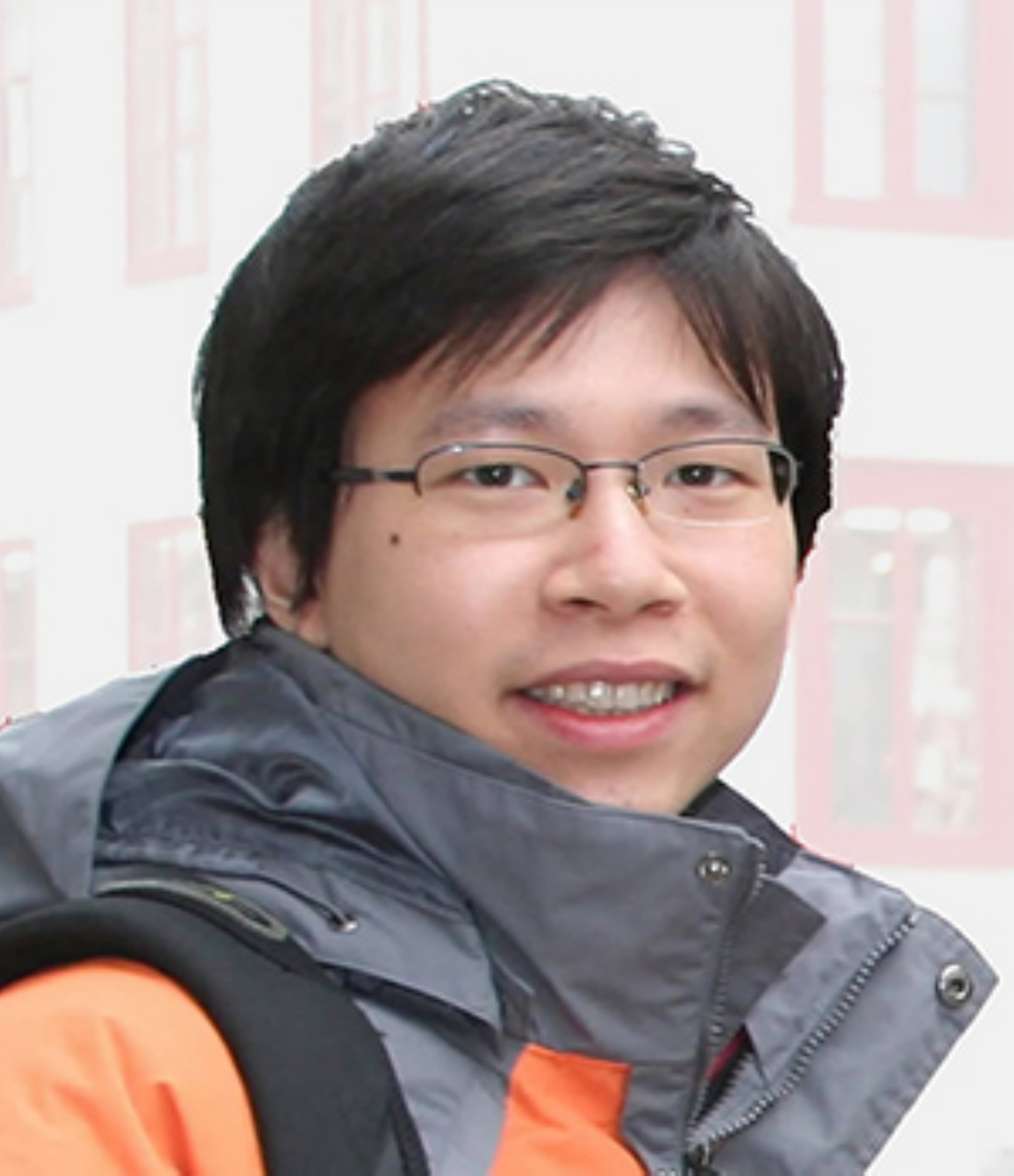}}]{Patrick P.K. Chan (M'04)} received the Ph.D. degree from Hong Kong Polytechnic University in 09. He is currently Associate Professor of School of Computer Science and Engineering in South China University of Technology, China. His current research interests include pattern recognition, adversarial learning, and multiple classifier systems. Dr. Chan is a member of the governing boards of IEEE SMC Society 14-16. He is also the Chairman of IEEE SMCS Hong Kong Chapter 14-15 and the counselor of IEEE Student Branch in South China University of Technology.
\end{IEEEbiography}

\vspace{-1.2 cm}

\begin{IEEEbiography}
[{\includegraphics[width=1in,height =1.25in,clip,keepaspectratio]{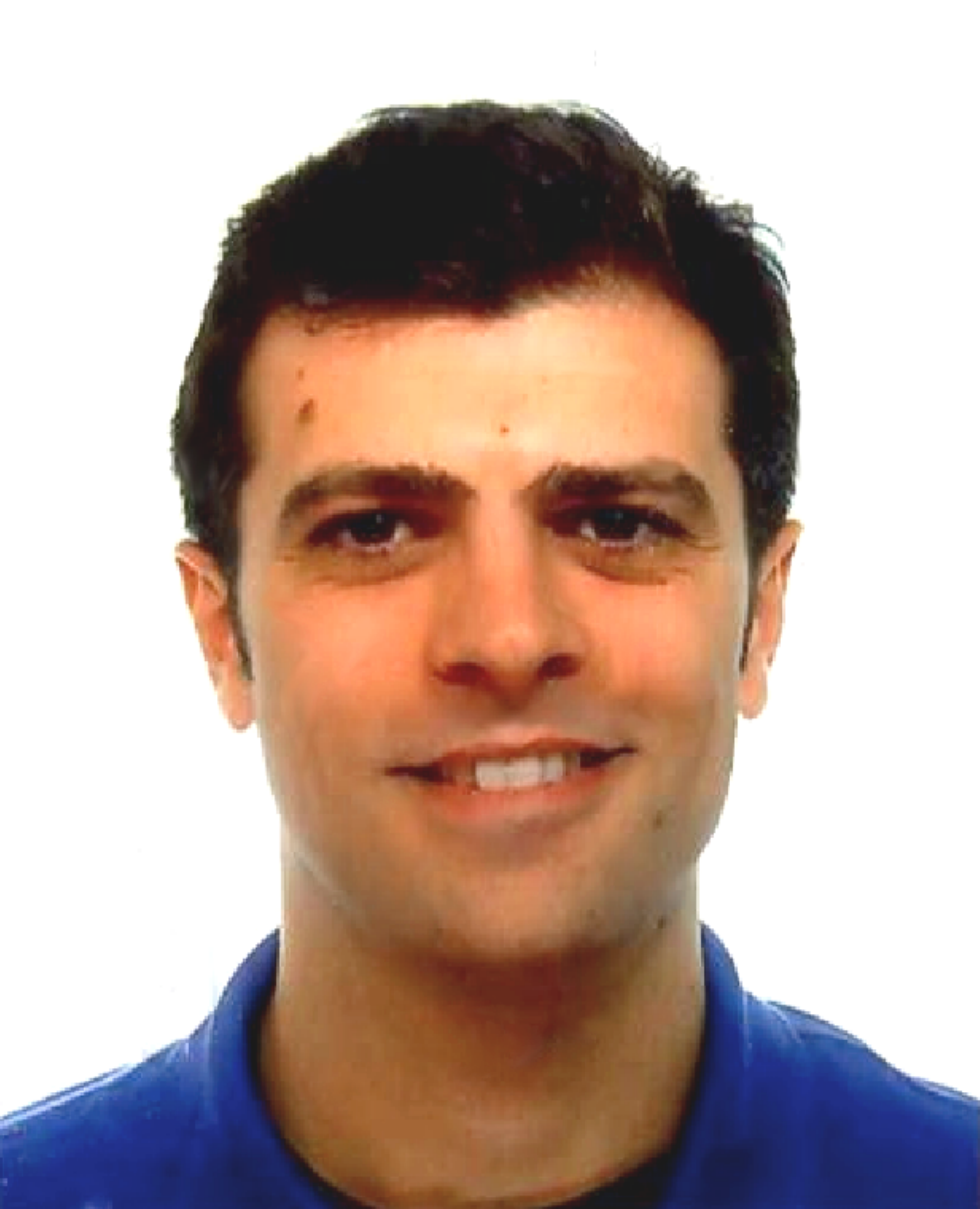}}]{Battista Biggio (M'07)} received the M.Sc. degree (Hons.) in electronic engineering and the Ph.D. degree in electronic engineering and computer science from the University of Cagliari, Italy, in 2006 and 2010. Since 2007, he has been with the Department of Electrical and Electronic Engineering, University of Cagliari, where he is currently a post-doctoral researcher. In 2011, he visited the University of T¨¹bingen, Germany, and worked on the security of machine learning to training data poisoning. His research interests include secure machine learning, multiple classifier systems, kernel methods, biometrics and computer security. Dr. Biggio serves as a reviewer for several international conferences and journals. He is a member of the IEEE and of the IAPR.
\end{IEEEbiography}

\vspace{-1.2 cm}

\begin{IEEEbiography}
[{\includegraphics[width=1in,height =1.25in,clip,keepaspectratio]{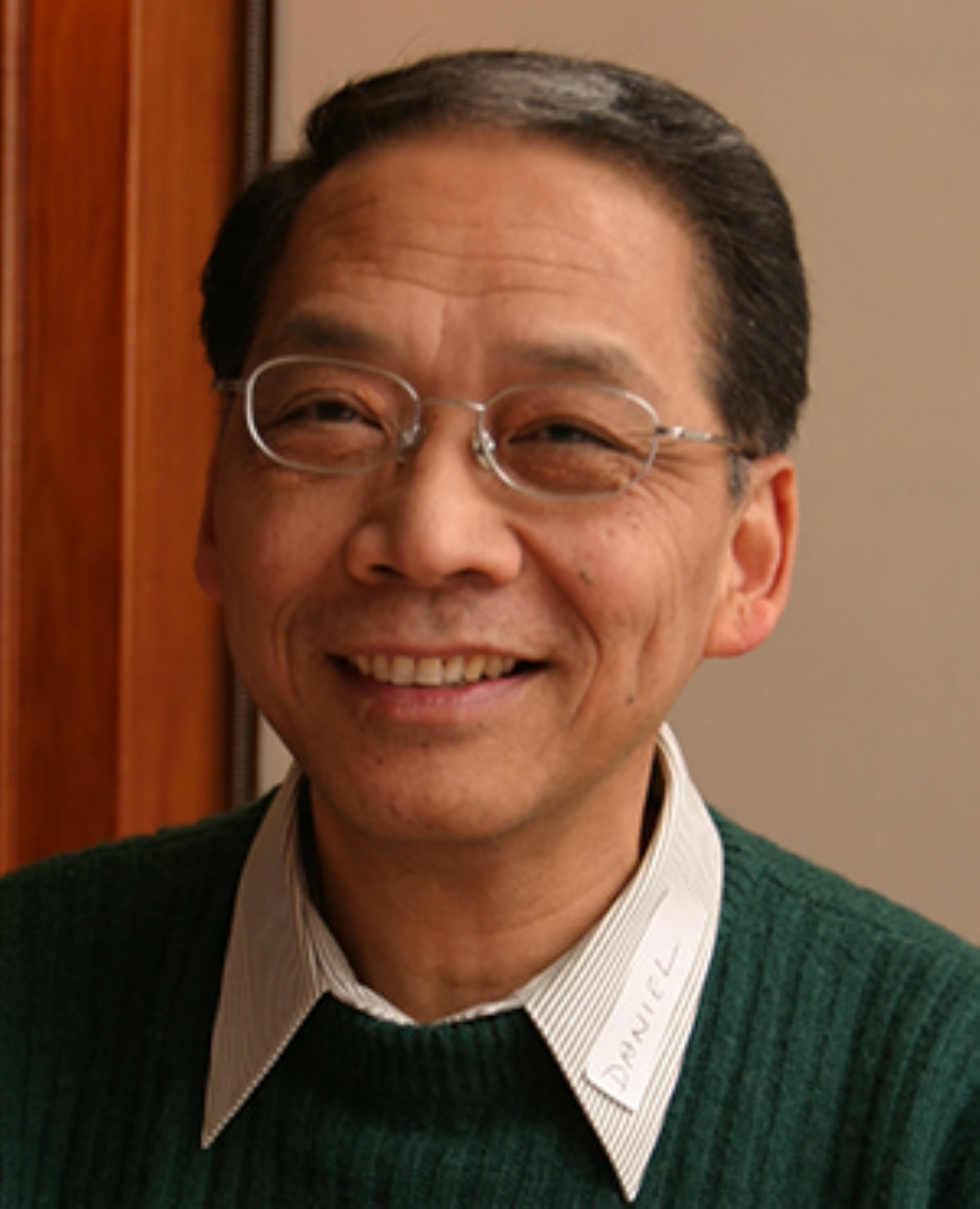}}]{Daniel S. Yeung (F'04)} received his Ph.D. from Case Western Reserve University, USA. He was a faculty at Rochester Institute of Technology from 74-80. In the next ten years he held industrial and business positions in USA. In 89 he joined City Polytechnic of Hong Kong as an Associate Head/Principal Lecturer at the Department of Computer Science. Then he served as the founding Head and Chair Professor of the Department of Computing at The Hong Kong Polytechnic University until his retirement at 06. Currently he is a Visiting Professor in the School of Computer Science and Engineering, South China University of Technology. Dr. Yeung is a fellow of the IEEE and served as the President of IEEE SMC Society in 08-09. His current research interests include neural-network sensitivity analysis, large scale data retrieval problem and cyber security.

\end{IEEEbiography}

\vspace{-1.2 cm}

\begin{IEEEbiography}[{\includegraphics[width=1in,height =1.25in,clip,keepaspectratio]{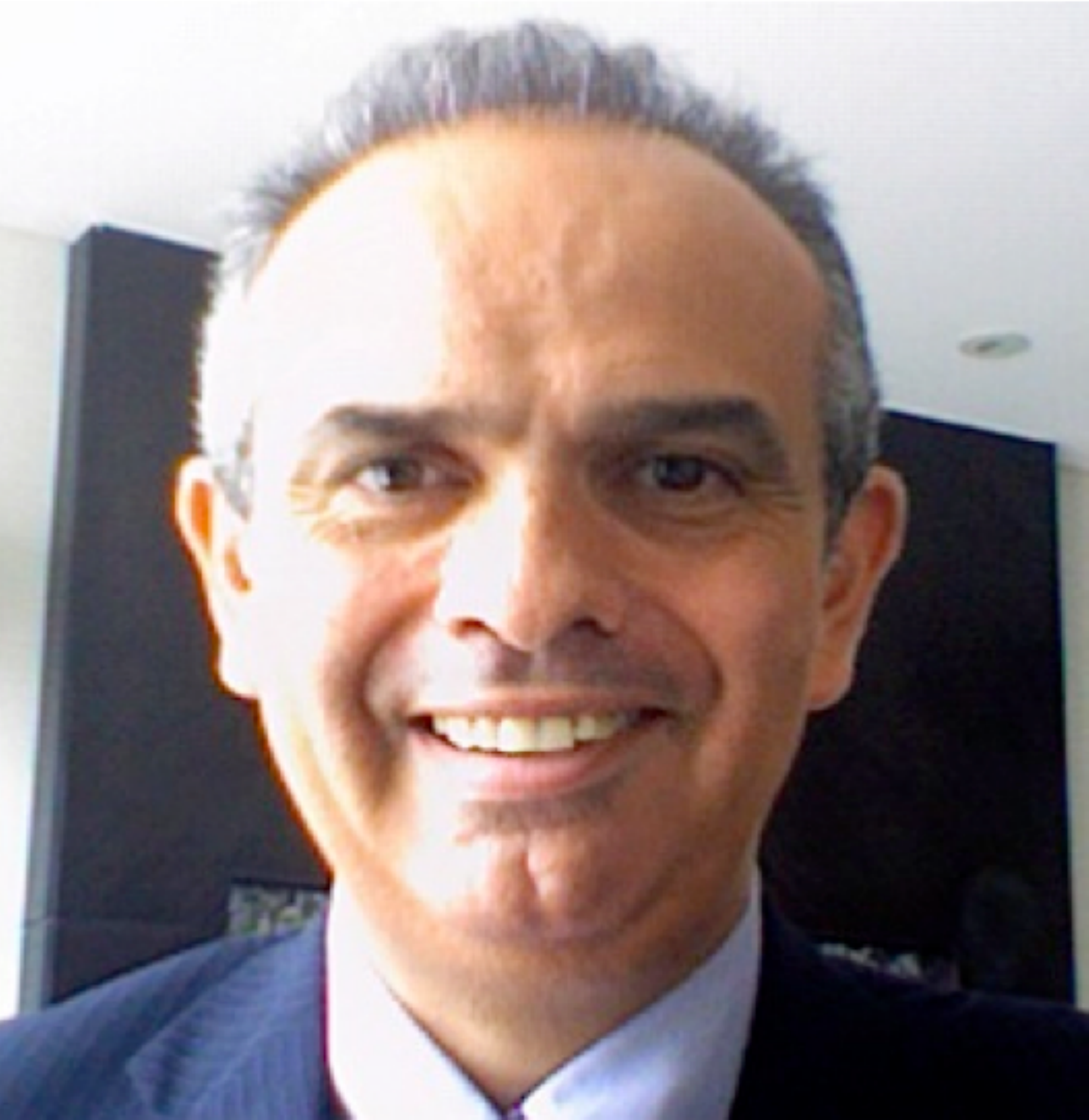}}]{Fabio Roli (F'12)}  received his Ph.D. in Electronic Eng. from the Univ. of Genoa, Italy. He was a research group member of the Univ. of Genoa (88-94). He was adjunct professor at the University of Trento (93-94). In 95, he joined the Dept. of Electrical and Electronic Eng. of the Univ. of Cagliari, where he is now professor of Computer Eng. and head of the research laboratory on pattern recognition and applications. His research activity is focused on the design of pattern recognition systems and their applications. He was a very active organizer of int'l conferences and workshops, and established the popular workshop series on multiple classifier systems. Dr. Roli is Fellow of the IEEE and of the Int'l Association for Pattern Recognition.

\end{IEEEbiography}

\end{document}